\documentclass{article}

\usepackage{microtype}
\usepackage{graphicx}
\usepackage{subfigure}
\usepackage{booktabs} 
\usepackage{amsmath}
\usepackage{comment}

\usepackage{amsmath,amsfonts,bm}

\def\eqref#1{equation~\ref{#1}}

\def\1{\bm{1}}

\DeclareMathAlphabet{\mathsfit}{\encodingdefault}{\sfdefault}{m}{sl}
\SetMathAlphabet{\mathsfit}{bold}{\encodingdefault}{\sfdefault}{bx}{n}

\newcommand{\R}{\mathbb{R}}

\DeclareMathOperator*{\argmax}{arg\,max}

\usepackage{mathtools}
\newcommand{\twobyone}[2]{\begin{bmatrix}#1 \\ #2 \end{bmatrix}}

\newcommand{\fun}[3]{\ensuremath{#1\colon #2\mapsto #3}}
\newcommand{\SO}{\ensuremath{\mathbf{SO}}}

\usepackage{hyperref}

\usepackage[accepted]{icml2021}

\usepackage{nicefrac}
\usepackage[capitalize]{cleveref}
\usepackage{siunitx}
\usepackage{multirow}

\usepackage[dvipsnames,table]{xcolor}
\usepackage{etoolbox}
\robustify\cellcolor
\newcommand{\bgl}{\cellcolor[HTML]{DDDDDD}}
\newcommand{\bgd}{\cellcolor[HTML]{BBBBBB}}
\renewcommand{\arraystretch}{1.2} 

\begin{document}

\twocolumn[
\icmltitle{Implicit-PDF: Non-Parametric Representation of Probability Distributions on the Rotation Manifold}
\icmlsetsymbol{equal}{*}

\begin{icmlauthorlist}
\icmlauthor{Kieran Murphy}{equal,goo}
\icmlauthor{Carlos Esteves}{equal,goo}
\icmlauthor{Varun Jampani}{goo}
\icmlauthor{Srikumar Ramalingam}{goo}
\icmlauthor{Ameesh Makadia}{goo}
\end{icmlauthorlist}

\icmlaffiliation{goo}{Google Research, New York, NY, USA}

\icmlcorrespondingauthor{}{implicitpdf@gmail.com}
\icmlkeywords{}

\vskip 0.3in
]

\printAffiliationsAndNotice{\icmlEqualContribution}

\begin{abstract}
Single image pose estimation is a fundamental problem in many vision and robotics tasks, and existing deep learning approaches suffer
by not completely modeling and handling: i) uncertainty about the predictions, and ii) symmetric objects with multiple (sometimes infinite) correct poses. 
To this end, we introduce a method to estimate arbitrary, non-parametric distributions on SO(3). 
Our key idea is to represent the distributions implicitly, with a neural network that estimates the probability given the input image and a candidate pose. 
Grid sampling or gradient ascent can be used to find the most likely pose, but it is also possible to evaluate the probability at any pose, enabling reasoning about symmetries and uncertainty. 
This is the most general way of representing distributions on manifolds, and to showcase the rich expressive power, we introduce a dataset of challenging symmetric and nearly-symmetric objects. 
We require no supervision on pose uncertainty -- the model trains only with a single pose per example. 
Nonetheless, our implicit model is highly expressive to handle complex distributions over 3D poses, while still obtaining accurate pose estimation on standard non-ambiguous environments, achieving state-of-the-art performance on Pascal3D+ and ModelNet10-SO(3) benchmarks. 
Code, data, and visualizations may be found at \href{https://implicit-pdf.github.io}{implicit-pdf.github.io}.

\end{abstract}

\section{Introduction}
\begin{figure*}[tp]
    \centering
    \includegraphics[width=\linewidth]{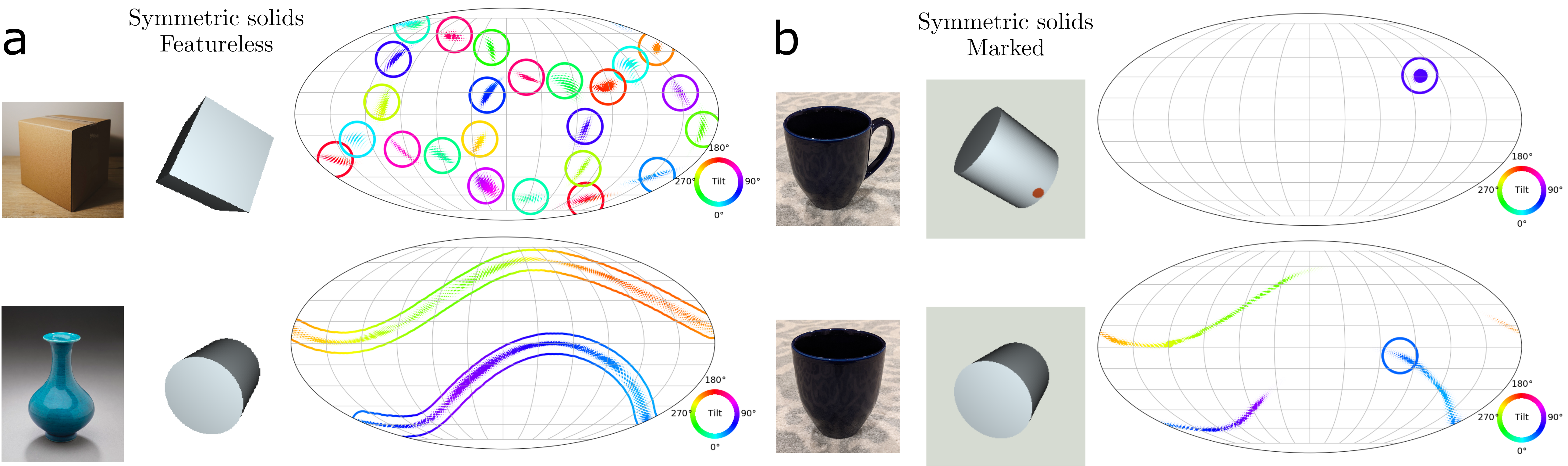}
    \caption{We introduce a method to predict arbitrary distributions over the rotation manifold. This is particularly useful for pose estimation of symmetric and nearly symmetric objects, since output distributions can include both uncertainty on the estimation and the symmetries of the object. \emph{\textbf{a-top:}} The cube has 24 symmetries, which are represented by 24 points on \SO(3), and all modes are correctly inferred by our model. \emph{\textbf{a-bottom:}} The cylinder has a continuous symmetry around one axis, which traces a cycle on \SO(3). It also has a discrete 2-fold symmetry (a ``flip''), so the distribution is represented as two cycles. The true pose distribution for the vase depicted on the left would trace a single cycle on \SO(3) since it does not have a flip symmetry. \emph{\textbf{b:}} This cylinder has a mark that uniquely identifies its pose, when visible (top). When the mark is not visible (bottom), our model correctly distributes the probability over poses where the mark is invisible. This example is analogous to a coffee cup when the handle is not visible. The resulting intricate distribution cannot be easily approximated with usual unimodal or mixture distributions on \SO(3), but is easily handled by our IPDF model. \emph{Visualization:} Points with non-negligible probability are displayed as dots on the sphere according to their first canonical axis, colored according to the rotation about that axis. The ground truth (used for evaluation only, not training) is shown as a solid outline. Refer to \cref{sec:method_visualization} for more details.}
\label{fig:teaser}
\end{figure*}

There is a growing realization in deep learning that bestowing a network with the ability to express uncertainty is universally beneficial and of crucial importance to systems where safety and interpretability are primary concerns~\cite{leibig2017leveraging,han2007uncertainties,ching2018opportunities}.  A quintessential example is the task of 3D pose estimation -- pose estimation is both a vital ingredient in many real-world robotics and computer vision applications where propagating uncertainty can facilitate complex downstream reasoning~\cite{mcallister2017concrete}, as well as an inherently ambiguous problem due to the abundant approximate and exact symmetries in our 3D world.

Many everyday objects possess symmetries such as the box or vase depicted in \cref{fig:teaser} (a). 
It is tempting to formulate a model of uncertainty that precisely mirrors the pose ambiguities of such  shapes; however it becomes immediately evident that such an approach is not scalable, as it is unrealistic to enumerate or characterize all sources of pose uncertainty. Even in a simple scenario such as a coffee mug with self-occlusion, the pose uncertainty manifests as a complex distribution over 3D orientations, as in \cref{fig:teaser} (b).

This paper addresses two long-standing and open challenges in pose estimation (a) \textit{what is the most general representation for expressing arbitrary pose distributions, including the challenging ones arising from symmetrical and near-symmetrical objects, in a neural network} and (b) \textit{how do we effectively train the model in typical scenarios where the supervision is a single 3D pose per observation} (as in Pascal3D+~\cite{pascal3d}, ObjectNet3D~\cite{objectnet}, ModelNet10-SO(3)~\cite{s2reg}), i.e.\ without supervision on the distribution, or priors on the symmetries.

To this end, we propose an \textit{implicit} representation for non-parametric probability distributions over the rotation manifold \SO(3) (we refer to our model as implicit-PDF, or IPDF for short). Such an implicit representation can be parameterized with a neural network and successfully trained with straightforward sampling strategies -- uniform or even random querying of the implicit function is sufficient to reconstruct the unnormalized distribution and approximate the normalizing term.
For inference, in addition to reconstructing the full probability distribution we can combine the sampling strategy with gradient ascent to make pose predictions at arbitrary (continuous) resolution. 
The use of a non-parametric distribution, while being simple, offers maximal expressivity for arbitrary densities and poses arising from symmetrical and near symmetrical 3D objects. The simplicity of our approach is in stark contrast to commonly used parametric distributions on \SO(3) that require complicated approximations for computing the normalizing term and further are not flexible enough to fit complex distributions accurately~\cite{gilitschenski2019deep,deng2020deep,mohlin2020probabilistic}.  Our primary contributions are
\begin{itemize}
\setlength\itemsep{0.2em}
\item \textit{Implicit}-PDF, a novel approach for modeling non-parametric distributions on the rotation manifold. Our implicit representation can be applied to realistic challenging pose estimation problems where uncertainty can arise from approximate or exact symmetries, self-occlusion, and noise.  We propose different sampling strategies which allow us to both efficiently reconstruct full distributions on \SO(3) as well as generate multiple pose candidates with continuous precision.

\item SYMSOL, a new dataset with inherent ambiguities for analyzing pose estimation with uncertainty. The dataset contains shapes with high order of symmetry, as well as nearly-symmetric shapes, that challenge probabilistic approaches to accurately learn complex pose distributions. When possible, objects are paired with their ground truth ``symmetry maps'', which allows quantitative evaluation of predicted distributions.
\end{itemize}
Our IPDF method is extensively evaluated on the new SYMSOL dataset as well as traditional pose estimation benchmarks. To aid our analysis, we develop a novel visualization method for distributions on \SO(3) that provides an intuitive way to qualitatively assess predicted distributions. Through evaluation of predicted distributions and poses, we obtain a broad assessment of our method: IPDF is the only technique that can consistently accurately recover the complex pose uncertainty distributions arising from a high degree of symmetry or self-occlusion, while being supervised by only a single pose per example. Further, while IPDF has the expressive power to model non-trivial distributions, it does not sacrifice in ability to predict poses in non-ambiguous situations and reaches state of the art performance with the usual metrics on many categories of Pascal3D+~\cite{pascal3d} and ModelNet10-SO(3)~\cite{s2reg}.

\section{Related work}
Symmetries are plentiful in our natural and human-made worlds, and so it is not surprising there is a history in computer vision of exploiting strong priors or assumptions on shape or texture symmetry to recover 3D structure from a single image~\cite{poggio1992recognition,hong2004symmetry,rothwell1993extracting}. However, among the more recent machine learning approaches for pose estimation, symmetries are treated as nuisances and strategies have been developed to utilize symmetry annotations at training.  With known symmetries at training, a canonical normalization of rotation space unambiguously resolves each set of equivalent rotations to a single one, allowing training to proceed as in single-valued regression~\cite{Pitteri2019}. In \citet{corona2018pose}, manually annotated symmetries on 3D shapes are required to jointly learn image embedding and classification of the object's symmetry order. Learning representations that cover a few specific symmetry classes is considered in~\citet{saxena09icra}.

In contrast to these works, \citet{Sundermeyer2019} make pose or symmetry supervision unnecessary by using a denoising autoencoder to %
isolate pose information. Neither~\citet{Sundermeyer2019} nor~\citet{corona2018pose} directly predict pose,
and thus require comparing against many rendered images of the same exact object for pose inference. In a similar vein, \citet{okorn2020learning} use a learned comparison against a dictionary of images to construct a histogram over poses. \citet{Fox-RSS-19} propose a particle filter framework for 6D object pose tracking, where each particle  represents a discrete distribution over \SO(3) with 191K bins. Similar to the previously mentioned works, this discrete rotation likelihood is estimated by codebook matching and an autoencoder is trained to generate the codes.

As noted earlier, symmetries are not the only source of pose uncertainty. Aiming to utilize more flexible representations, a recent direction of work has looked to directional statistics~\cite{mardiabook} to consider parameteric probability distributions. Regression to the parameters of a von Mises distribution over (Euler) angles~\cite{prokudin2018deep}, as well as regression to the Bingham~\cite{peretroukhin_so3_2020,deng2020deep,gilitschenski2019deep} and Matrix Fisher distributions~\cite{mohlin2020probabilistic} over \SO(3) have been proposed.  Since it is preferable to train these probabilistic models with a likelihood loss, the distribution's normalizing term must be computed, which is itself a challenge (it is a hypergeometric function of a matrix argument for Bingham and Matrix Fisher distributions).  \citet{gilitschenski2019deep} and \citet{deng2020deep} approximate this function and gradient via interpolation in a lookup table, \citet{mohlin2020probabilistic} use a hand-crafted approximation scheme to compute the gradient, and \citet{peretroukhin_so3_2020} simply forgo the likelihood loss.  In the simplest setting these models are unimodal, and thus ill equipped to deal with non-trivial distributions. To this end, \citet{prokudin2018deep}, \citet{gilitschenski2019deep}, and \citet{deng2020deep} propose using multimodal mixture distributions. One challenge to training the mixtures is avoiding mode collapse, for which a winner-take-all strategy can be used~\cite{deng2020deep}.  
An alternative to the mixture models is to directly predict multiple pose hypotheses~\cite{Manhardt_2019_ICCV}, but this does not share any of the benefits of a probabilistic representation.

Bayesian deep learning provides a general framework to reason about model uncertainty, and in \citet{kendall2016modelling} test time dropout~\cite{gal2016dropout} was used to approximate Bayesian inference for camera relocalization. Inference with random dropout applied to the trained model is used to generate Monte Carlo pose samples, and thus this approach does not offer a way to estimate the density at arbitrary poses (sampling large numbers of poses would also be impractical).

An alternative framework for representing arbitrary complex distributions is Normalizing Flows~\cite{rezende2015variational}.  In principle, the reparameterization trick for Lie groups introduced in \citet{falorsi2019reparameterizing} allows for constructing flows to the Lie algebra of \SO(3).
\citet{rezende2020normalizing} develop normalizing flows for compact and connected differentiable manifolds, however it is still unclear how to effectively construct flows on non-Euclidean manifolds, and so far there has been little evidence of a successful application to realistic problems at the complexity of learning arbitrary distributions on \SO(3).

The technical design choices of our implicit pose model are inspired by the very successful implicit shape \cite{MeschederONNG19} and scene \cite{mildenhall2020nerf} representations, which can represent detailed geometry with a multilayer perceptron that takes low-dimensional position and/or directions as inputs.  

We introduce the details of our approach next.

\section{Methods}
The method centers upon a multilayer perceptron (MLP) which implicitly represents probability distributions over \SO(3). The input to the MLP is a pair comprising a rotation and a visual representation of an image obtained using a standard feature extractor such as a residual network; the output is an unnormalized log probability. Roughly speaking, we construct the distribution for a given image by populating the space of rotations with such queries, and then normalizing the probabilities. This procedure is highly parallelizable and efficient (see Supp.\ for time ablations).  In the following we provide details for the key ingredients of our method.

\subsection{Formalism}
\label{sec:method_formalism}
Our goal is, given an input $x \in \mathcal{X}$ (for example, an image),
to obtain a conditional probability distribution $\fun{p(\cdot | x)}{\SO(3)}{\R^+}$,
that represents the pose of $x$.
We achieve this by training a neural network to estimate the
unnormalized joint log probability function \fun{f}{\mathcal{X} \times \SO(3)}{\R}.
Let $\alpha$ be the normalization term such that $p(x, R) = \alpha \exp(f(x, R))$,
where $p$ is the joint distribution.
The computation of $\alpha$ is infeasible, requiring integration over $\mathcal{X}$.
From the product rule, $p(R|x) = p(x, R) / p(x)$.
We estimate $p(x)$ by marginalizing over \SO(3),
and since \SO(3) is low-dimensional, we approximate the integral with a discrete sum as follows,
\begin{align}
   p(x) &= \int_{R\in\SO(3)}p(x, R)\,dR \nonumber\\
        &= \alpha \int_{R\in\SO(3)}\exp(f(x, R))\,dR \nonumber\\
        &\approx \alpha \sum_i^N \exp(f(x, R_i))V,
\end{align}
where the $\{R_i\}$ are centers of an equivolumetric partitioning of \SO(3) with $N$ partitions of volume $V = \nicefrac{\pi^2}{N}$.
(see \cref{sec:method_sampling} for details). Now $\alpha$ cancels out in the expression for $p(R|x)$, giving
\begin{align}
  \label{eq:conditional}
  p(R|x) \approx \frac{1}{V}\frac{\exp(f(x, R))}{\sum_i^N \exp(f(x, R_i))},
\end{align}
where all the RHS terms are obtained from querying the neural network.

During training, the model receives pairs of inputs $x$ and corresponding ground truth $R$,
and the objective is to maximize $p(R|x)$.
See \cref{sec:method_loss} for details.

\paragraph{Inference -- single pose.}
To make a single pose prediction, we solve
\begin{align}
  R_x^* = \argmax_{R \in \SO(3)} f(x, R),
\end{align}
with gradient ascent, since $f$ is differentiable.
The initial guess comes from evaluating a grid $\{R_i\}$.
Since the domain of this optimization problem is \SO(3),
we project the values back into the manifold after each gradient ascent step.

\paragraph{Inference -- full distribution.} Alternatively, we may want to predict a full probability distribution.
In this case ${p(R_i|x)}$ is evaluated over the \SO(3) equivolumetric partition $\{R_i\}$. This representation allows us to reason about uncertainty and observe complex patterns of symmetries and near-symmetries.

Our method can estimate intricate distributions on the manifold
without direct supervision of such distributions.
By learning to maximize the likelihood of a single ground truth pose per object over a dataset,
with no prior knowledge of each object's symmetries,
appropriate patterns expressing symmetries and uncertainty naturally emerge in our model's outputs,
as shown in \cref{fig:teaser}.

\subsection{Network}
\label{sec:method_network}
Inspired by recent breakthroughs in implicit shape and scene representations \cite{MeschederONNG19,ParkFSNL19,SitzmannZW19},
we adopt a multilayer perceptron (MLP) to implicitly represent the pose distribution. 
Differently from most implicit models, we train a single model to represent the pose of any instance of multiple categories, so an input descriptor (e.g.\ pre-trained CNN features for image inputs) is also fed to the MLP, which we produce with a pre-trained ResNet~\cite{resnet}.  
Most implicit representation methods for shapes and scenes
take a position in Euclidean space and/or a viewing direction as inputs. In our case, we take an arbitrary 3D rotation,
so we must revisit the longstanding question of how to represent rotations~\cite{levinson20neurips}. 
We found it best to use a $3\times 3$ rotation matrix to avoid discontinuities present in other representations~\cite{saxena09icra}. Following \citet{mildenhall2020nerf}, we found positionally encoding each element of the input to be beneficial. See the supplement for ablative studies on these design choices.

\subsection{Loss}
\label{sec:method_loss}
We train our model by minimizing the predicted negative log-likelihood of the (single) ground truth pose. This requires normalizing the output distribution, which we approximate by evaluating \cref{eq:conditional} using the method described in \cref{sec:method_sampling} to obtain an equivolumetric grid over \SO(3), in which case the normalization is straightforward. During training, we rotate the grid such that $R_0$ coincides with the ground truth. Then, we evaluate $p(R_0 | x)$ as in \cref{eq:conditional}, and the loss is simply
\begin{align}
  \mathcal{L}(x, R_0) = -\log(p(R_0 | x))
\end{align}
We noticed that the method is robust enough to be trained without an equivolumetric grid; evaluating \cref{eq:conditional} for randomly sampled $R_i \in \SO(3)$, provided that one of them coincides with the ground truth, works similarly well. The equivolumetric partition is still required during inference for accurate representation of the probabilities.

\subsection{Sampling the rotation manifold}
\label{sec:method_sampling}
Training and producing an estimate of the most likely pose does not require precise normalization of the  probabilities predicted by the network.  
However, when the distribution is the object of interest (e.g.\ an accurate distribution will be used in a downstream task), we can normalize by evaluating on a grid of points with equal volume in \SO(3) and approximating the distribution as a histogram.

We employ a method of generating equivolumetric grids developed by~\citet{yershova2010generating}, which uses as its starting point the HEALPix method of generating equal area grids on the 2-sphere~\cite{Gorski_2005}.  
A useful property of this sampling is that it is generated hierarchically, permitting multi-resolution sampling if desired.

\begin{figure}[t]
    \centering
    \includegraphics[width=0.47\linewidth]{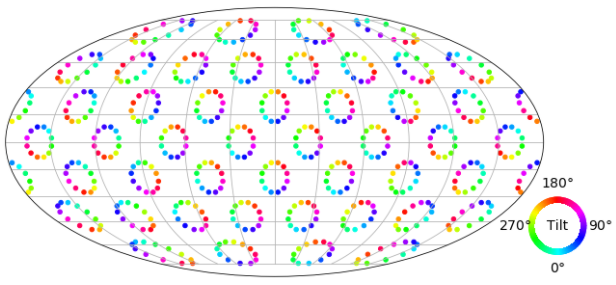}
    \includegraphics[width=0.47\linewidth]{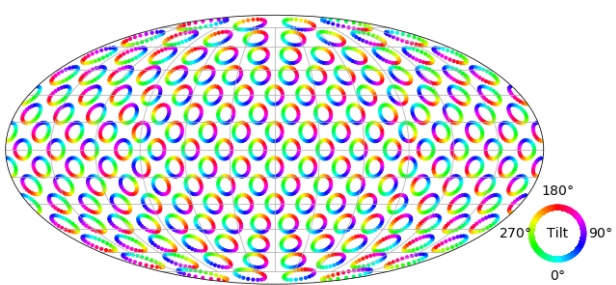}
\caption{\textbf{Equivolumetric grid on \SO(3).} In order to normalize the output distribution, we sample unnormalized densities on an equivolumetric grid following~\citet{yershova2010generating}.  This iterative method starts with HEALPix~\cite{Gorski_2005} which generates equal-area grids hierarchically on the sphere. Left: a grid with 576 samples, right: 4608 samples.}
    \label{fig:grid_so3}
\end{figure}

The Hopf fibration is leveraged to cover \SO(3) by threading a great circle through each point on the surface of a 2-sphere.
The grids are generated recursively from a starting seed of 72 points, and grow by a factor of eight each iteration.  
Figure~\ref{fig:grid_so3} shows grids after one and two subdivisions.
For evaluation, we use the grid after 5 subdivisions, with a little more than two million points.

\begin{figure*}[ht]
    \centering
    \includegraphics[width=1\linewidth]{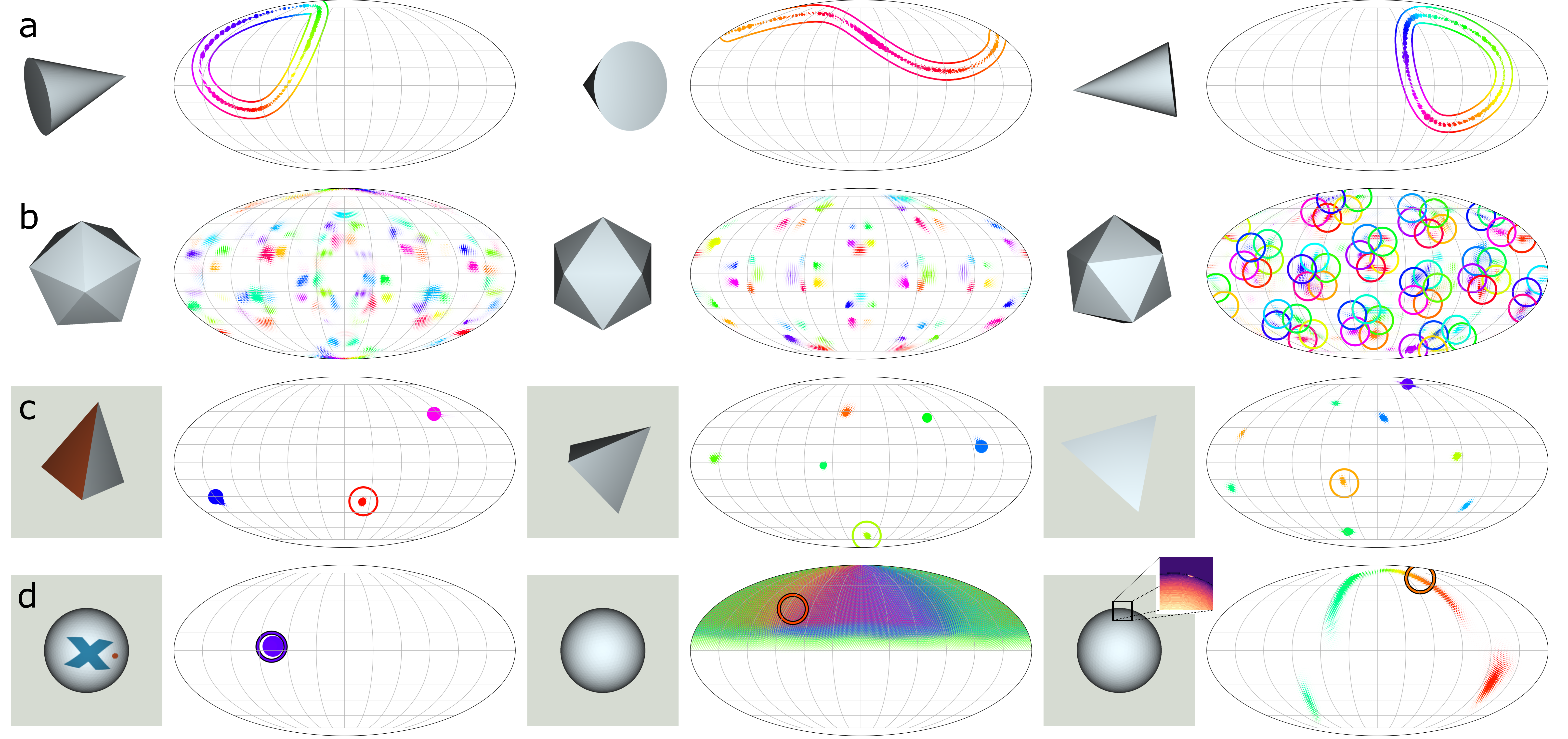}
\caption{\textbf{IPDF predicted distributions for SYMSOL.} \textbf{(a)} The cone has one great circle of equivalent orientations under symmetry.  \textbf{(b)} The 60 modes of icosahedral symmetry would be exceedingly difficult for a mixture density network based approach, but IPDF can get quite close (we omit the ground truths from the left and middle visualizations for clarity).  \textbf{(c)} The marked tetrahedron (``tetX'') has one red face.  When it is visible, the 12-fold tetrahedral symmetry reduces to only three equivalent rotations.  With less information about the location of the red face, more orientations are possible: 6 when two white faces are visible (middle) and 9 when only one white face is visible (right).
\textbf{(d)} The orientation of the marked sphere (``sphereX'') is unambiguous when both markings are visible (left).  When they are not (middle), all orientations with the markings on the hidden side of the sphere are possible. When only a portion of the markings are visible (right; inset is a magnification showing several pixels of the X are visible), the IPDF distribution captures the partial information.
}
\label{fig:symsol_combo}
\end{figure*}

\subsection{Visualization}
\label{sec:method_visualization}
We introduce a novel method to display distributions over \SO(3).  
A common approach to visualizing such distributions is via multiple marginal distributions,
e.g. over each of the three canonical axes~\cite{lee_global_symplectic,mohlin2020probabilistic}.
This is in general incomplete as it is not able to fully specify the joint distribution.

In order to show the full joint distribution, we display the entire space of rotations with the help of the Hopf fibration.
With this method, we project a great circle of points on \SO(3) to each point on the 2-sphere, and then use the color wheel to indicate the location on the great circle.
More intuitively, we may view each point on the 2-sphere as the direction of a canonical z-axis, and the color indicates the tilt angle about that axis.
To represent probability density, we vary the size of the points on the plot.
Finally, we display the surface of the 2-sphere using the Mollweide projection.

As the method projects to a lower dimensional space, there are limitations arising from occlusions, but also a freedom in the projection axis which allows finding more or less informative views.
The visualization benefits from relatively sparse distributions where much of the space has negligible probability.
We did not find this to be limiting in practice: even the 60 modes of a distribution expressing icosahedral symmetry are readily resolved
(Fig.~\ref{fig:symsol_combo}b).
\begin{table*}[t]
\caption{Distribution estimation on SYMSOL I and II.
    We report the average log likelihood on both parts of the SYMSOL dataset, as a measure for how well the multiple equivalent ground truth orientations are represented by the output distribution.
    For reference, a minimally informative uniform distribution over \SO(3) has an average log likelihood of -2.29. 
    IPDF's expressivity allows it to more accurately represent the complicated pose distributions across all of the shapes.
    A separate model was trained for each shape for all baselines and for all of SYMSOL II, but only a single IPDF model was trained on all five shapes of SYMSOL I.
    }
    \centering
  {
    \begin{tabular}{@{}l
      S[table-format=2.2,table-figures-decimal=2,table-auto-round]
      S[table-format=1.2,table-figures-decimal=2,table-auto-round]S[table-format=1.2,table-figures-decimal=2,table-auto-round]S[table-format=1.2,table-figures-decimal=2,table-auto-round]
      S[table-format=1.2,table-figures-decimal=2,table-auto-round]S[table-format=1.2,table-figures-decimal=2,table-auto-round]
      c
      S[table-format=2.2,table-figures-decimal=2,table-auto-round]      
      S[table-format=1.2,table-figures-decimal=2,table-auto-round]S[table-format=1.2,table-figures-decimal=2,table-auto-round]S[table-format=1.2,table-figures-decimal=2,table-auto-round]
      }
                                                & \multicolumn{6}{c}{SYMSOL I (log likelihood $\uparrow$)}                           &           & \multicolumn{4}{c}{SYMSOL II (log likelihood $\uparrow$)}                                                                       \\
      \cmidrule{2-7}
      \cmidrule{9-12}
                                                & {avg.}                                                 & {cone}    & {cyl.}     & {tet.}    & {cube}     & {ico.}     &  & {avg.}    & {sphX}    & {cylO}    & {tetX}    \\
      \midrule
                  \citet{deng2020deep}          & -1.482                                                 & 0.16      & -0.95      & \bgl 0.27 & -4.44      & -2.45      &  & 2.573     & 1.12      & 2.99      & \bgl 3.61 \\
                  \citet{gilitschenski2019deep} & \bgl -0.43                                             & \bgl 3.84 & \bgl  0.88 & -2.29     & -2.29      & \bgl -2.29 &  & \bgl 3.7  & \bgl 3.32 & \bgl 4.88 & 2.90      \\
                  \citet{prokudin2018deep}      & -1.874                                                 & -3.34     & -1.28      & -1.86     & \bgl -0.50 & -2.39      &  & 0.483     & -4.19     & 4.16      & 1.48      \\
                  IPDF (Ours)                   & \bgd 4.1                                               & \bgd 4.45 & \bgd 4.26  & \bgd 5.70 & \bgd 4.81  & \bgd 1.28  &  & \bgd 7.57 & \bgd 7.30 & \bgd 6.91 & \bgd 8.49 \\
      \bottomrule
    \end{tabular}
  }
\label{tab:symsol_loglik}
\end{table*}

\subsection{Evaluation metrics}
\label{sec:method_metrics}
The appropriateness of different metrics depends on the nature of predictions (a probability distribution or a set of values) and on the state of knowledge of the ground truth.

\noindent
\textbf{Prediction as a distribution: Log likelihood} 
In the most general perspective, ground truth annotations accompanying an image are \textit{observations} from an unknown distribution which incorporates symmetry, ambiguity, and human error involved in the process of annotation.
The task of evaluation is a comparison between two distributions given samples from one, for which likelihood is standard~\cite{gan,ClausetPowerlaws2009,okorn2020learning,gilitschenski2019deep}.
We report the log likelihood averaged over test set annotations, $\mathbb{E}_{x \sim p(x), R \sim p_\text{GT}(R|x)}[\text{log}\ p(R|x)]$.
Importantly, the average log likelihood is invariant to whether one ground truth annotation is available or a set of all equivalent annotations.

\noindent
\textbf{Prediction as a distribution: Spread} 
When a complete set of equivalent ground truth values is known (e.g. a value for each equivalent rotation under symmetry), the expected angular deviation to any of the ground truth values is $\mathbb{E}_{R \sim p(R|x)}[\min_{R^\prime \in \{R_\text{GT}\}} d(R,R^\prime)]$ and 
\fun{d}{\SO(3) \times \SO(3)}{\R^+} is the geodesic distance between rotations.
This measure has been referred to as the Mean Absolute Angular Deviation (MAAD)  \cite{prokudin2018deep,gilitschenski2019deep}, and encapsulates both the deviation from the ground truths and the uncertainty around them.

\noindent
\textbf{Prediction as a finite set: precision}
The most common evaluation scenario in pose estimation tasks is a one-to-one comparison between a single-valued prediction and a ground truth annotation.
However, in general, both the prediction and ground truth may be multi-valued, though often only one of the ground truths is available for evaluation.
To compensate, sometimes symmetries are implicitly imposed on the entire dataset by reporting flip-invariant metrics~\cite{suwajanakorn2018discovery,esteves19icml}.
These metrics evaluate precision, where a prediction need only be close to one of the ground truths to score well.
Usually, the median angular error and accuracy at some angular threshold $\theta$ are  reported in this setting.

\textbf{Prediction as a finite set: recall}
We can also evaluate the coverage of multiple ground truths given multiple predictions, indicating recall.
We employ a simple method of clustering by connected components to extract multiple predictions from an output distribution, and rank by probability mass, to return top-\textit{k} recall metrics;
median error and accuracy at $\theta$ are evaluated in this setting.
When $k=1$ and the ground truth is unique, these coincide with the precision metrics.
See the supplement for extended discussion.

\section{Experiments}

\subsection{Datasets}
To highlight the strengths of our method,
we put it to the test on a range of challenging pose estimation datasets.%

First, we introduce a new dataset (\textbf{SYMSOL I}) of images rendered around simple symmetric solids.
It includes images of platonic solids (tetrahedron, cube, icosahedron)
and surfaces of revolution (cone, cylinder),
with 100,000 renderings of each shape from poses sampled uniformly at random from \SO(3).
Each image is paired with its ground truth symmetries
(the set of rotations of the source object that would not change the image),
which are easily derived for these shapes.
As would be the case in most practical situations, 
where symmetries are not known and/or only approximate, we
use such annotations only for evaluation and not for training.
Access to the full set of equivalent rotations opens new avenues of evaluating model performance rarely possible with pose estimation datasets.

While the textureless solids generate a challenging variety of distributions, 
they can still be approximated with mixtures of simple unimodal distributions such as the Bingham~\cite{deng2020deep,gilitschenski2019deep}.
We go one step further and break the symmetry of objects by texturing with small markers (\textbf{SYMSOL II}).
When the marker is visible, the pose distribution is no longer ambiguous and collapses given the extra information.
When the marker is not visible, only a subspace of the symmetric rotations for the textureless shape are possible.

For example, consider a textureless sphere.
Its pose distribution is uniform -- rotations will not change the input image.
Now suppose we mark this sphere with a small arrow.
If the arrow is visible, the pose distribution collapses to an impulse.
If the arrow is not visible, the distribution is no longer uniform,
since about half of the space of possible rotations can now be eliminated.
This distribution cannot be easily approximated by mixtures of unimodals.

\textbf{SYMSOL II} objects include a sphere
marked with a small letter ``X'' capped with a dot to break flip symmetry when visible (sphX),
a tetrahedron with one red and three white faces (tetX),
and a cylinder marked with a small filled off-centered circle (cylO).
We render 100,000 images for each.

The two SYMSOL datasets test expressiveness, but the solids are relatively simple and the dataset does not require generalization to unseen objects.
\textbf{ModelNet10-SO(3)} was introduced by \citet{s2reg} to study pose estimation
on rendered images of CAD models from ModelNet10~\cite{wu20153d}.
As in SYMSOL, the rotations of the objects cover all of \SO(3) and therefore present
a difficulty for methods that rely on particular rotation formats such as Euler angles~\cite{s2reg,prokudin2018deep}.

The \textbf{Pascal3D+} dataset~\cite{pascal3d} is a popular benchmark for pose estimation on real images,
consisting of twelve categories of objects.
Though some of the categories contain instances with symmetries (e.g. bottle and table),
the ground truth annotations have generally been disambiguated and restricted to subsets of \SO(3).
This allows methods which regress to a single pose to perform competitively~\cite{s2reg}.
Nevertheless, the dataset is a challenging test on real images.

Finally, we %
evaluate on \textbf{T-LESS}~\cite{hodan2017tless}, 
consisting of texture-less industrial parts with various discrete and continuous approximate symmetries.
As in~\citet{gilitschenski2019deep}, we use the Kinect RGB single-object images, tight-cropped and color-normalized.
Although the objects are nearly symmetric,
their symmetry-breaking features are visible in most instances.
Nonetheless, it serves as a useful benchmark to compare distribution
metrics with \citet{gilitschenski2019deep}.

We find that IPDF proves competitive across the board.

\subsection{Baselines}
We compare to several recent works which parameterize distributions on $\SO(3)$ for the purpose of pose estimation.
\citet{gilitschenski2019deep} and \citet{deng2020deep} output the parameters for mixtures of Bingham distributions and interpolate from a large lookup table to compute the normalization constant.
\citet{mohlin2020probabilistic} output the parameters for a unimodal matrix Fisher distribution and similarly employ an approximation scheme to compute the normalization constant.
\citet{prokudin2018deep} decompose \SO(3) into the product of three independent distributions over Euler angles, with the capability for multimodality through an `infinite mixture' approach.
Finally we compare to the spherical regression work of~\citet{s2reg}, which directly regresses to Euler angles, to highlight the comparative advantages of distribution-based methods.
We quote reported values and run publicly released code when values are unavailable.
See Supplemental Material for additional details.

\begin{table}[t]
\caption{ModelNet10-SO(3) accuracy and median angle error.
    Metrics are averaged over categories.
    Our model can output pose candidates, so we also evaluate top-$k$ metrics,
    which are more robust to the lack of symmetry annotations in this dataset.
    See Supplementary Material for the complete table with per-category metrics.
  }
  \centering
  \resizebox{0.48\textwidth}{!}{
    \begin{tabular}{@{}l
      ccc
      }
                                      & {Acc@15\textdegree  $\uparrow$} & {Acc@30\textdegree  $\uparrow$} & \shortstack{Med.\ ($^\circ$) $\downarrow$} \\
      \midrule
      \citet{s2reg}                   & 0.496                           & 0.658                           & 28.7                                       \\      
      \citet{deng2020deep}            & 0.562                           & 0.694                           & 32.6                                       \\
      \citet{prokudin2018deep}        & 0.456                           & 0.528                           & 49.3                                       \\
      \citet{mohlin2020probabilistic} & \bgl 0.693                      & \bgd 0.757                      & \bgd 17.1                                  \\    
      IPDF (ours)                     & \bgd 0.719                      & \bgl 0.735                      & \bgl 21.5                                  \\
      \midrule
      IPDF (ours), top-2              & 0.868                           & 0.888                           & 4.9                                        \\
      IPDF (ours), top-4              & 0.904                           & 0.926                           & 4.8                                        \\
      \bottomrule
    \end{tabular}
    }
  \label{tab:m10so3}
\end{table}

\subsection{SYMSOL I: symmetric solids}
\label{sec:symsolid}
We report the average log likelihood in \Cref{tab:symsol_loglik}, 
and the gap between IPDF and the baselines is stark.
The average log likelihood indicates how successful the prediction is at distributing probability mass around \textit{all} of the ground truths. 
The expressivity afforded by our method allows it to capture both the continuous and discrete symmetries present in the dataset.
As the order of the symmetry increases from 12 for the tetrahedron, to 24 for the cube, and finally 60 for the icosahedron,
the baselines struggle and tend to perform at same level as a minimally informative (uniform) distribution over \SO(3).
The difference between IPDF and the baselines in \Cref{tab:symsol_loglik} is further cemented by the fact that a single IPDF model was trained on all five shapes while the baselines were allowed a separate model per shape.
Interestingly, while the winner-take-all strategy of~\citet{deng2020deep} enabled training with more Bingham modes than~\citet{gilitschenski2019deep}, it seems to have hindered the ability to faithfully represent the continuous symmetries of the cone and cylinder, as suggested by the relative performance of these methods.

\begin{figure}[t]
    \centering
    \includegraphics[width=1\linewidth]{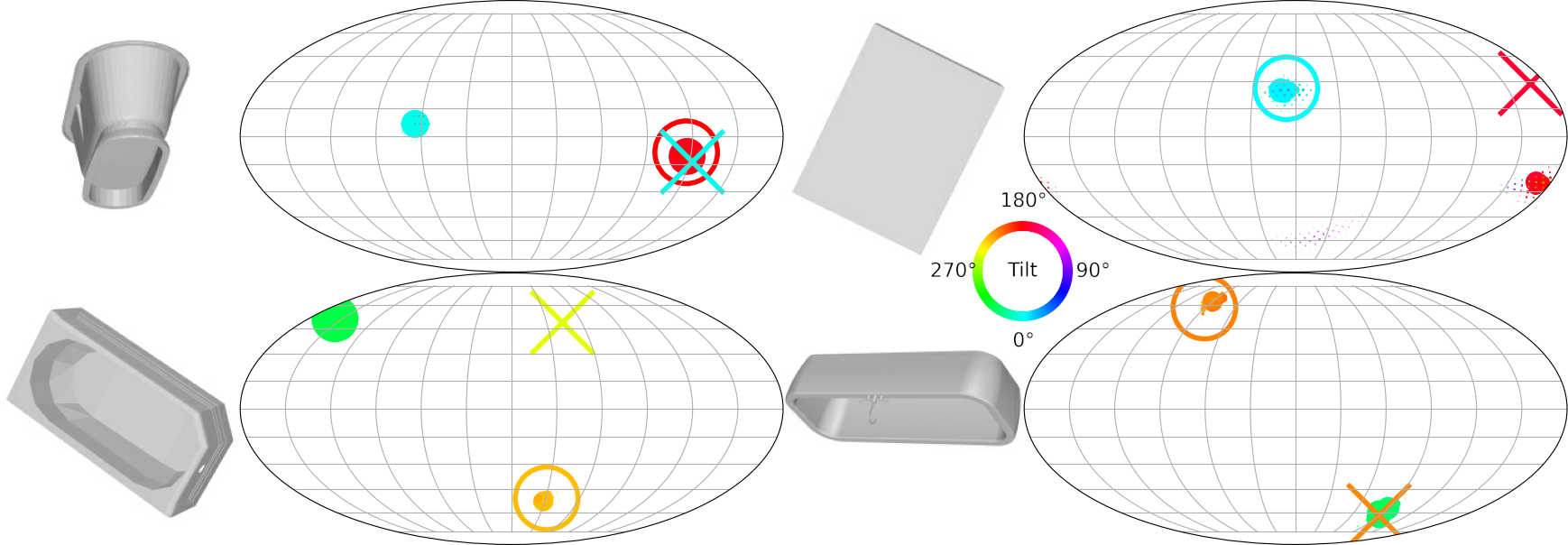}
    \caption{Bathtubs may have exact or approximate 2-fold symmetries around one or more axes.
      We show our predicted probabilities as solid disks,
      the ground truth as circles, and the predictions of \citet{s2reg} as crosses.
      Our model assigns high probabilities to all symmetries,
      while the regression method ends up far from every symmetry mode
      (note the difference in position and color between circles and crosses).}
    \label{fig:bathtubs}
\end{figure}

\begin{table*}[t]
\caption{Results on a standard pose estimation benchmark, Pascal3D+. As is common, we show accuracy at 30$^\circ$ (top) and median error in degrees (bottom), for each category and also averaged over categories.  Our IPDF is at or near state-of-the-art on many categories. \textbf{\ddag} The results for \citet{s2reg} and \citet{mohlin2020probabilistic} differ from their published numbers. For \citet{s2reg}, published errors are known to be incorrectly scaled by a $\sqrt{2}$ factor, and \citet{mohlin2020probabilistic} evaluates on a non-standard test set. See Supplemental for details.}
  \label{tab:pascal}
  \centering
  \resizebox{0.99\textwidth}{!}{
    \begin{tabular}{@{}ll@{\hskip 0.3in}
      c@{\hskip 0.3in}
      cccccc
      cccccc
      }
                                         &                                      & {avg.}     & {aero}  & {bike}      & {boat}    & {bottle}     & {bus}  & {car}       & {chair} & {table}     & {mbike}    & {sofa} & {train}  & {tv} \\
      \midrule
      \multirow{6}{*}{Acc@30\textdegree  $\uparrow$} & \ddag\citet{s2reg}           & 0.819 & 0.82 & 0.77 & 0.55 & 0.93 & 0.95 & 0.94 & 0.85 & 0.61 & 0.80 & \bgl 0.95 & \bgd 0.83 & 0.82     \\      
                                         & \ddag\citet{mohlin2020probabilistic} & 0.825 & \bgd 0.90 & \bgd 0.85 & 0.57 & 0.94 & 0.95 & \bgd 0.96 & 0.78 & 0.62 & 0.87 & 0.85 & 0.77 & 0.84 \\
                                         & \citet{prokudin2018deep} & \bgl 0.838 & \bgl 0.89 &  0.83 & 0.46 & \bgd 0.96 & 0.93 & 0.90 & 0.80 & \bgl 0.76 & \bgd 0.90 & 0.90 & \bgl 0.82 & \bgd 0.91 \\
                                         & \citet{Tulsiani_2015_CVPR} & 0.808 & 0.81  & 0.77  & \bgl 0.59 & 0.93 & \bgd 0.98 & 0.89 & 0.80 & 0.62 & \bgl 0.88 & 0.82 & 0.80 & 0.80 \\
                                         & \citet{mahendranmixed} & \bgd 0.859 & 0.87 & 0.81 & \bgd 0.64 & \bgd 0.96 & \bgl 0.97 & \bgl 0.95 & \bgd 0.92 & 0.67 & 0.85 & \bgd 0.97 & \bgl 0.82 & \bgl 0.88 \\
                                         & IPDF (Ours)                 & 0.837 & 0.81 & \bgd 0.85 & 0.56 & 0.93 & 0.95 & 0.94 & \bgl 0.87 & \bgd 0.78 & 0.85 & 0.88 & 0.78 & 0.86  \\
                                         
      \midrule
      \multirow{6}{*}{\shortstack{Median\\error ($^\circ$)  $\downarrow$}}
                                         & \ddag\citet{s2reg}            & 13.0 & 13.0 & 16.4 & 29.1 & 10.3 & 4.8 & 6.8 & 11.6 & 12.0 & 17.1 & 12.3 & 8.6 & 14.3       \\
                                           & \ddag\citet{mohlin2020probabilistic} & 11.5 & 10.1 & 15.6 & 24.3 & 7.8 & 3.3 & 5.3 & 13.5 & 12.5 & \bgl 12.9 & 13.8 & 7.4 & \bgd 11.7\\
                                           & \citet{prokudin2018deep} & 12.2 & \bgl 9.7 & 15.5 & 45.6 & \bgd 5.4 & \bgd 2.9 & \bgd 4.5 & 13.1 & 12.6 & \bgd 11.8 & \bgd 9.1 & \bgd 4.3 & 12.0 \\
                                           & \citet{Tulsiani_2015_CVPR} & 13.6 & 13.8 & 17.7 & \bgl 21.3 & 12.9 & 5.8 & 9.1 & 14.8 & 15.2 & 14.7 & 13.7 & 8.7 & 15.4 \\
                                           & \citet{mahendranmixed} & \bgd 10.1 & \bgd 8.5 & \bgl 14.8 & \bgd 20.5 & \bgl 7.0 & \bgl 3.1 & \bgl 5.1 & \bgd 9.3 & \bgl 11.3 & 14.2 & 10.2 & \bgl 5.6 & \bgd 11.7 \\
                                           & IPDF (Ours)                        & \bgl 10.3 & 10.8 & \bgd 12.9 & 23.4 & 8.8 & 3.4 & 5.3 & \bgl 10.0 & \bgd 7.3 & 13.6 & \bgl 9.5  & 6.4  & 12.3 \\
      
      \bottomrule
    \end{tabular}
  }
  \end{table*}

\subsection{SYMSOL II: nearly-symmetric solids}
\label{sec:symsolii}

When trained on the solids with distinguishing features which are visible only from a subset of orientations, 
IPDF is far ahead of the baselines (Table~\ref{tab:symsol_loglik}).
The prediction serves as a sort of `belief state', with the flexibility of being unconstrained by a particular parameterization of the distribution.
The marked cylinder in the right half of Figure~\ref{fig:teaser} displays this nicely.  
When the red marking is visible, the pose is well defined from the image and the network outputs a sharp peak at the correct, unambiguous location.
When the cylinder marking is not visible, there is irreducible ambiguity conveyed in the output with half of the full cylindrical symmetry shown in the left side of the figure.

The pose distribution of the marked tetrahedron in Figure~\ref{fig:symsol_combo}c takes a discrete form.
Depending on which faces are visible, a subset of the full 12-fold tetrahedral symmetry can be ruled out.
For example, with the one red face visible in the left subplot of Figure~\ref{fig:symsol_combo}c, there is nothing to distinguish the three remaining faces, and the implicit distribution reflects this state with three modes.

Figure~\ref{fig:symsol_combo}d show the IPDF prediction for various views of the marked sphere. 
When the marking is not visible at all (middle subplot), the half of \SO(3) where the marking faces the camera can be ruled out; IPDF assigns zero probability to half of the space.
When only a portion of the marking is visible (right subplot), IPDF yields a nontrivial distribution with an intermediate level of ambiguity, capturing the partial information contained in the image.

\begin{figure}[ht]
    \centering
    \includegraphics[width=0.9\columnwidth]{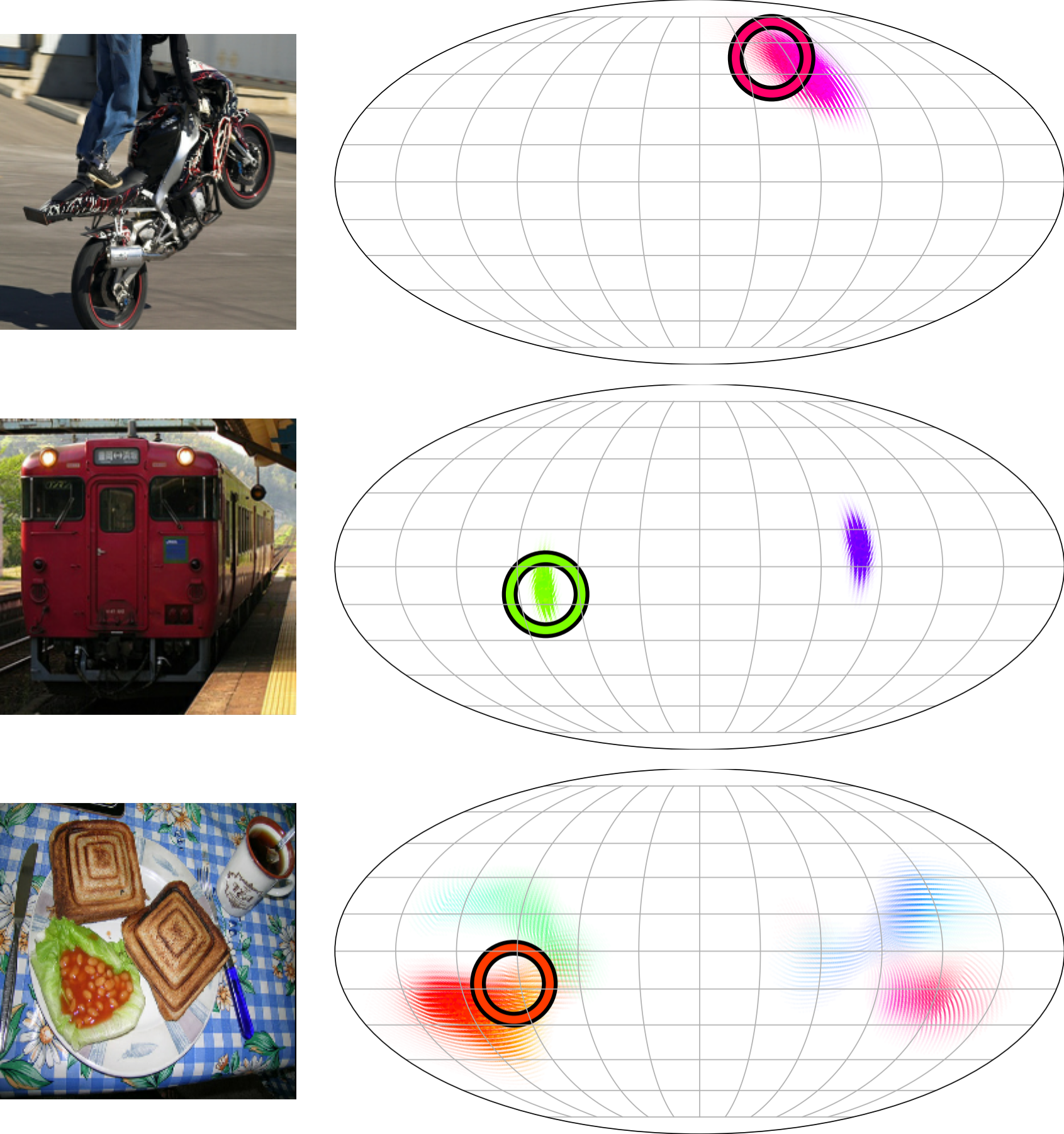}
\caption{\textbf{IPDF predicted distributions on Pascal3D+.} 
We display a sampling of IPDF pose predictions to highlight the richness of information contained in the full distribution output, as compared to a single pose estimate.
Uncertainty regions and multi-modal predictions are freely expressed, owing to the non-parametric nature of IPDF.
}
\label{fig:pascal_tri}
\end{figure}

\subsection{ModelNet10-SO(3)}
\label{sec:m10}
Unimodal methods perform poorly on categories with rotational symmetries such as \emph{bathtub},
\emph{desk} and \emph{table} (see the supplementary material for complete per-category results).
When trained with a single ground truth pose selected randomly from among multiple distinct rotations, 
these methods tend to split the difference and predict a rotation equidistant from all equivalent possibilities.
The most extreme example of this behavior is the \emph{bathtub} category,
which contains instances with approximate or exact two-fold symmetry
around one or more axes (see \cref{fig:bathtubs}).
With two modes of symmetry separated by 180$^\circ$, 
the outputs tend to be 90$^\circ$ away from each mode.
We observe this behavior in \citet{s2reg,mohlin2020probabilistic}. 

Since our model can easily represent any kind of symmetry,
it does not suffer from this problem, as illustrated in \cref{fig:bathtubs}.
The predicted distribution captures the symmetry of the object but returns only
one of the possibilities during inference.
This is penalized by metrics that rely on a single ground truth,
since picking the mode that is not annotated results in an 180$^\circ$ error,
while picking the midpoint between two modes (which is far from both)
results in a 90$^\circ$ error.
Since some \emph{bathtub} instances have two-fold symmetries over more than one axis
(like the top-right of \cref{fig:bathtubs}),
our median error ends up closer to 180$^\circ$ when the symmetry annotation is incomplete,
which in turn significantly increases the average over all categories.
We observe the same for other multi-modal methods \cite{prokudin2018deep,deng2020deep}.

Our performance increases dramatically in the top-$k$ evaluation even for $k=2$ (see \cref{tab:m10so3}).
The ability to output pose candidates is an advantage of our model,
and is not possible for direct regression~\cite{s2reg}
or unimodal methods~\cite{mohlin2020probabilistic}.
While models based on mixtures of unimodal distributions could, in theory,
produce pose candidates, their current implementations~\cite{gilitschenski2019deep,deng2020deep} suffer from mode collapse and are constrained to a fixed number of modes.

\subsection{Pascal3D+}
In contrast to the full coverage of \SO(3) and the presence of symmetries and ambiguities in the SYMSOL and ModelNet10-SO(3) datasets,
Pascal3D+ serves as a check that pose estimation performance in the unambiguous case is not sacrificed.
In fact, as the results of Table~\ref{tab:pascal} show, IPDF performs as well as or better than the baselines which constitute a variety of methods to tackle the pose estimation problem.
The feat is remarkable given that our method was designed for maximal expressiveness and not for the single-prediction, single-ground truth scenario.
IPDF performance in terms of median angular error, while good, overlooks the wealth of information contained in the full predicted distribution.
Sample pose predictions are shown in Figure~\ref{fig:pascal_tri} and in the Supplemental; the distributions express uncertainty and category-level pose ambiguities.

\begin{table}[h]
\caption{Pose estimation on T-LESS.
LL is the log-likelihood, spread is the mean angular error, and Med. is the median angular error for single-valued predictions.
\citet{gilitschenski2019deep} underestimate its evaluation of spread, disregarding the dispersion.
}
\centering
\resizebox{0.48\textwidth}{!}{
  \begin{tabular}{@{}l %
    S[table-format=2.1,table-figures-decimal=1,table-auto-round]S[table-format=2.1,table-figures-decimal=1,table-auto-round]S[table-format=2.1,table-figures-decimal=1,table-auto-round]
    }
                                       & {LL $\uparrow$}      & {Spread ($^\circ$) $\downarrow$}    & {Med.\ ($^\circ$)  $\downarrow$} \\
    \midrule
    \citet{deng2020deep}          &  5.3 & 23.1      & 3.11           \\
    \citet{gilitschenski2019deep} & 6.88      & \bgd 3.44 & 2.65           \\
    \citet{prokudin2018deep} & \bgl 8.8 & 34.3 & \bgd 1.248 \\
    \citet{s2reg}                 & {-}       & {-}       & 2.61      \\
    IPDF (Ours)                     & \bgd 9.80 & \bgl 4.12 & \bgl 1.31      \\
    \bottomrule
  \end{tabular}
}
  \label{tab:tless}
\end{table}

\subsection{T-LESS}
The results of Table~\ref{tab:tless}, and specifically the success of the regression method of~\citet{s2reg},
show that approximate or exact symmetries are not an issue in the particular split of the T-LESS dataset used in~\citet{gilitschenski2019deep}.
All methods are able to achieve median angular errors of less than 4$^\circ$.
Among the methods which predict a probability distribution over pose, IPDF maximizes the average log likelihood and minimizes the spread,
when correctly factoring in the uncertainty into the metric evaluation.

\section{Conclusion}
In this work we have demonstrated the capacity of an implicit function to represent highly expressive,
non-parametric distributions on the rotation manifold.
It performs as well as or better than state of the art parameterized distribution methods, on standard pose estimation benchmarks where the ground truth is a single pose.
On the new and difficult SYMSOL dataset, the implicit method is far superior while being simple to implement as it does not require any onerous calculations of a normalization constant.
Particularly, we show in SYMSOL II that our method can represent distributions that cannot be approximated well by current mixture-based models.
See the Supplementary Material for additional visualizations, ablation studies and timing evaluations, extended  discussion about metrics, and implementation details.

\bibliography{references}
\bibliographystyle{icml2021}

\robustify\cellcolor
\renewcommand{\arraystretch}{1.2} 

\onecolumn

\newpage

\hrule

{\Large Supplemental Material for \textit{Implicit-PDF: Non-Parametric Representation of Probability Distributions on the Rotation Manifold}}

\setcounter{section}{0}
\setcounter{page}{1}
\setcounter{figure}{0}
\setcounter{table}{0}

\renewcommand{\thepage}{S\arabic{page}}
\renewcommand{\thesection}{S\arabic{section}}
\renewcommand{\thetable}{S\arabic{table}}
\renewcommand{\thefigure}{S\arabic{figure}}
\renewcommand{\figurename}{Supplemental Material, Figure}

\section{Additional IPDF predictions for objects from Pascal3D+}

\begin{figure}[h]
    \centering
    \includegraphics[width=1\linewidth]{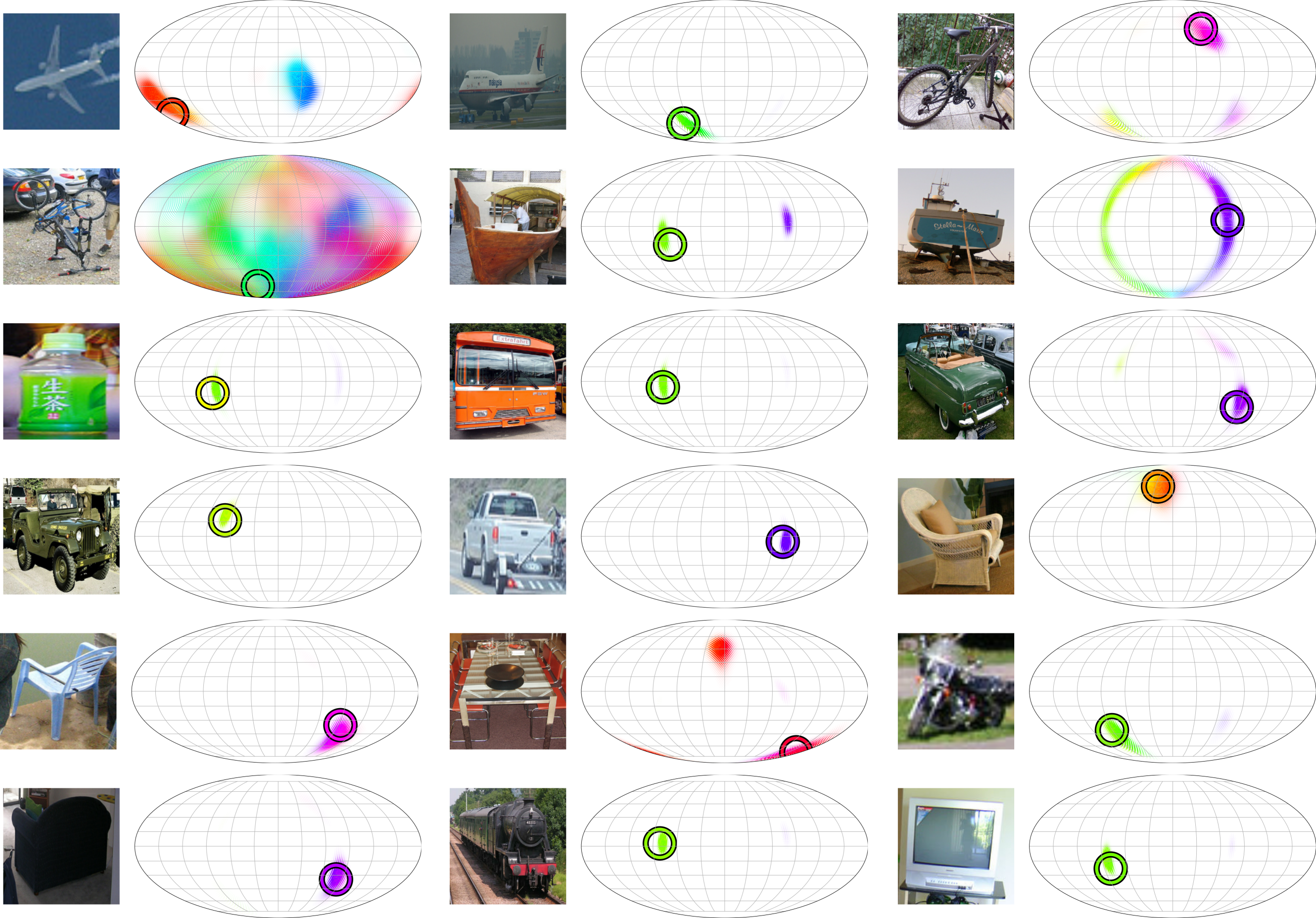}
    \caption{\textbf{Sample IPDF outputs on Pascal3D+ objects.}
    We visualize predictions by the IPDF model, trained on all twelve object categories, which yielded the results in Table 3 of the main text.
    The ground truth rotations are displayed as the colored open circles.
    }
    \label{fig:pascal_assortment}
\end{figure}

In Figure~\ref{fig:pascal_assortment} we show sample predictions from IPDF trained on the objects in Pascal3D+.
The network outputs much more information about the pose of the object in the image than can be expressed in a single estimate.
Even in the examples where the distribution is unimodal, and the pose is relatively unambiguous, IPDF provides rich information about the uncertainty around the most likely pose.
The expressivity of IPDF allows it to express category-level symmetries, which appear as multiple modes in the distributions above.
The most stand-out example in Figure~\ref{fig:pascal_assortment} is of the bicycle in the second row: the pose estimate of IPDF is incredibly uncertain, yet still there is information in the exclusion of certain regions of \SO(3) which have been `ruled out'.
The expressivity of IPDF allows an unprecedented level of information to be contained in the predicted pose distributions.

\section{Extension of IPDF beyond \SO(3)}
\begin{figure}[h]
    \centering
    \includegraphics[width=0.6\linewidth]{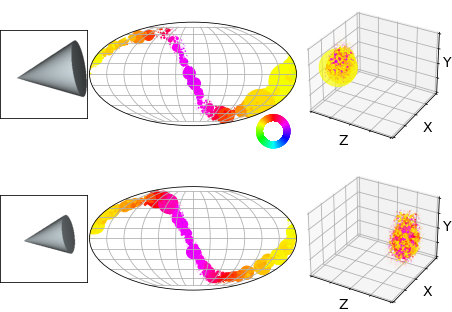}
    \caption{\textbf{Extension to 6DOF rotation+translation estimation.}
    We train IPDF on a modified SYMSOL I dataset, where the objects are also translated in space.
    Shown above are two images of a cone with the same orientation but shifted in space.
    We query the network over the full joint space of translations and rotations, and visualize the marginal distributions.
    Each point in rotation space has a corresponding point in translation space, and we color them the same to indicate as such.
    While uninformative in the above plots, this scheme of coloring allows nontrivial joint distributions to be expressed.
    }
\label{fig:supp_6dof}
\end{figure}

IPDF is not limited to probability distributions on \SO(3), which nevertheless served as a challenging and practical testing ground for the method.
With minor modifications, IPDF can be extended to the problem of pose with six degrees of freedom (6DOF): we append translation coordinates to the rotation query, and use $10{\times}$ more samples during training to adequately cover the full joint space. 
Normalizing the distributions is similarly straightforward, by querying over a product of Cartesian and HealPix-derived grids. 
Predicted distributions on modified images of SYMSOL are shown in Figure~\ref{fig:supp_6dof}. 
For two renderings of a cone from identical orientation but different translations, only the predicted distribution over translation differs between the two images.

\section{SYMSOL spread evaluation, compared to multimodal Bingham}

\begin{figure}[h]
    \centering
    \includegraphics[width=0.79\linewidth]{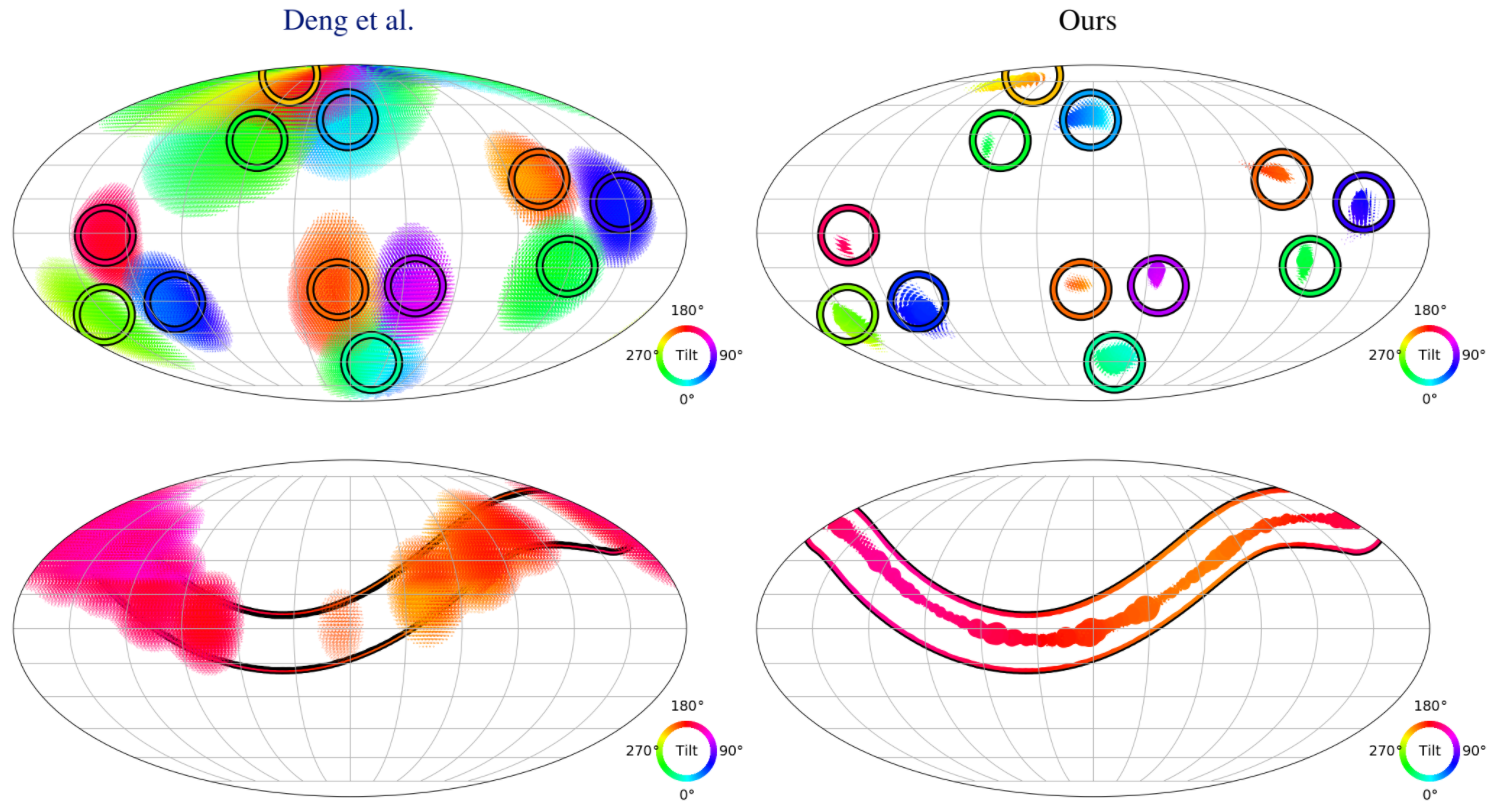}
    \caption{\textbf{Comparison of predicted distributions: tetrahedron and cone.}
    We show predicted pose distributions for a tetrahedron (top) and cone (bottom).  
    Displayed on the left is the method of \citet{deng2020deep}, which outputs parameters for a mixture of Bingham distributions.
    The right side shows IPDF.
    The predicted distributions from the implicit method are much more densely concentrated around the ground truth, providing a visual grounding for the significant difference in the spread values of Table~\ref{tab:supp_spread}.
    }
    \label{fig:supp_birdal_viz}
\end{figure}

\begin{table}[h]
  \caption{Spread estimation on SYMSOL.  This metric evaluates how closely the probability mass is centered on any of the equivalent ground truths.
  For this reason, we can only evaluate it on SYMSOL I, where all ground truths are known at test time.
  Values are in degrees.}
  \centering
  {
    \begin{tabular}{@{}l
      S[table-format=1.2,table-figures-decimal=2,table-auto-round]S[table-format=1.2,table-figures-decimal=2,table-auto-round]S[table-format=1.2,table-figures-decimal=2,table-auto-round]
      S[table-format=1.2,table-figures-decimal=2,table-auto-round]S[table-format=1.2,table-figures-decimal=2,table-auto-round]}
      \toprule
                                                     & {cone} & {cyl.}    & {tet.}    & {cube}    & {ico.}    \\
      \midrule
                  \citeauthor{deng2020deep}          &{10.1} & {15.2} & {16.7} & {40.7} & {29.5}  \\
                  Ours                               & \bgd {1.4} & \bgd {1.4} & \bgd {4.6} & \bgd {4.0} & \bgd {8.4}  \\
      \bottomrule
    \end{tabular}
  }
\label{tab:supp_spread}
\end{table}

We evaluate the spread metric on the SYMSOL I dataset, where the full set of ground truths is known at test time, for IPDF and the method of \citet{deng2020deep}.  The results are shown in Table~\ref{tab:supp_spread}.

The metric values, in degrees, show how well the implicit method is able to home in on the ground truths. 
For the cone and cylinder, the spread of probability mass away from the continuous rotational symmetry has a typical scale of just over one degree.

The predicted distributions in Figure~\ref{fig:supp_birdal_viz} for a tetrahedron and cone visually ground the values of Table~\ref{tab:supp_spread}.
Many of the individual unimodal Bingham components can be identified for the output distributions of \citet{deng2020deep}, highlighting the difficulty covering the great circle of ground truth rotations for the cone with only a limited number of unimodal distributions (bottom).
The spread around the ground truths for both shapes is significantly larger and more diffuse than for IPDF, shown on the right.

\section{Computational cost}
We evaluate the computational cost of our method by measuring the time it takes
to obtain the pose distribution for a single image,
which corresponds to the frequency it could run on real time.
The fair baseline here is the direct regression method of \citet{s2reg},
using the same ResNet-50 backbone and the same size of MLP.
The only difference is that while \citet{s2reg} only feeds the image descriptor to the MLP,
our model concatenates the descriptor to a number of query poses from a grid.

\Cref{tab:time} shows the results. When using the coarser grid, the performance overhead is negligible
with respect to the baseline.
This grid has approximately 5$^\circ$ between nearest neighbors, which might be enough for some applications.
When increased accuracy is required, our model can use more samples, trading speed for accuracy.
Note that the MLP operations are highly parallelizable on GPUs so
the processing time grows slower than linear with the grid size.
\begin{table*}[ht]
  \caption{Inference time evaluation. For our method, we measure the time needed to generate the normalized distribution over SO(3) given a single 224 $\times$ 224 image.
    The number of samples correspond to the HEALPix-SO(3) grids of levels 3, 4, and 5, respectively.
    The coarser grid has an average distance of approximately 5$^\circ$ between nearest neighbors.
    The processing time growth is slower than linear.
  }
  \label{tab:time}
  \centering
  {
    \begin{tabular}{@{}lcS[table-format=2.1,table-figures-decimal=1,table-auto-round]
      S[table-format=1.3,table-figures-decimal=3,table-auto-round]S[table-format=1.3,table-figures-decimal=3,table-auto-round]S[table-format=2.1,table-figures-decimal=1,table-auto-round]
      @{}
      }
      \toprule
       {Method}          & {Number of samples} & {frames/s $\downarrow$} & {Acc@15\textdegree  $\uparrow$} & {Acc@30\textdegree  $\uparrow$} & \shortstack{Med.\ ($^\circ$) $\downarrow$} \\
      \midrule
      \citeauthor{s2reg} & {-}                 & \bgd 18.2               & 0.5224                          & 0.6516                          & 38.2                                       \\
      Ours               & \SI{37}{k}          & \bgl 18.3               & \bgl 0.717                      & \bgl 0.735                      & 25.1                                  \\
      Ours               & \SI{295}{k}         & 9.13                    & \bgd 0.723                      & \bgd 0.738                      & \bgd 17.6                                  \\
      Ours               & \SI{2359}{k}        & 2.39                    & \bgd 0.723                      & \bgd 0.738                      & \bgl 18.7                                  \\            
      \bottomrule
    \end{tabular}
  }
\end{table*}

\section{Ablations}

\begin{figure}[htbp]
    \centering
    \includegraphics[width=1\linewidth]{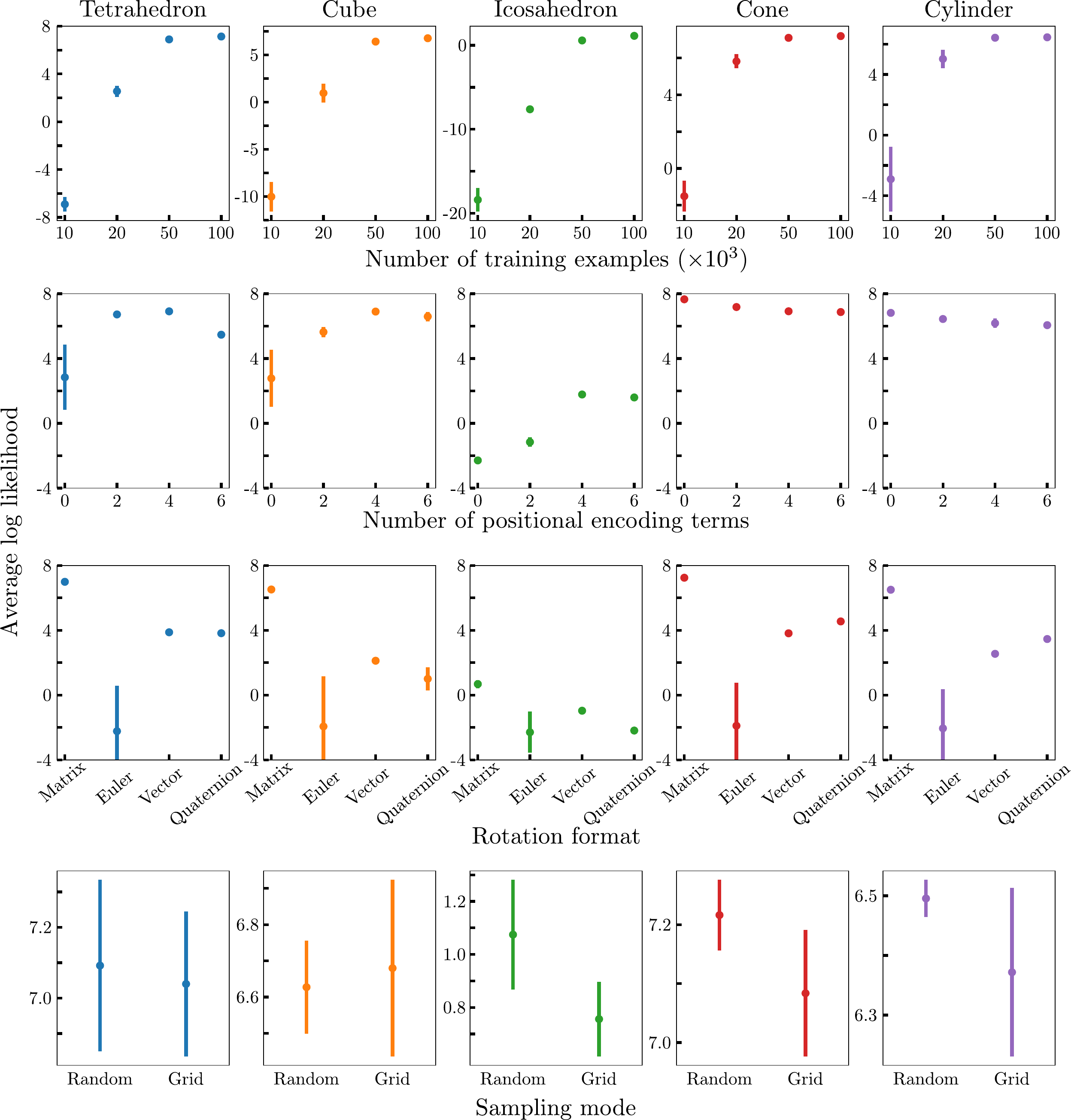}
    \caption{\textbf{Ablative studies.} We report the average log likelihood for the shapes of SYMSOL I with various aspects of the method ablated. 
    Error bars are the standard deviation over five networks trained with different random seeds.
    In the top row, we show the dependence on the size of the dataset, with performance leveling off after 50,000 images per shape.
    The subsequent row varies the positional encoding, with 0 positional encoding terms corresponding to no positional encoding at all: the flattened rotation matrix is the query rotation.
    The third row examines the role of the rotation format when querying the MLP (before positional encoding is applied).
    The final row shows that, during training, inexact normalization arising from the queries being randomly sampled over \SO(3) leads to roughly equivalent performance as the proper normalization from using the equivolumetric grid as the query points.
    Note that evaluation makes use of an equivolumetric grid in both cases, to calculate the log likelihood.
    }
\label{fig:supp_ablations}
\end{figure}

In Figure~\ref{fig:supp_ablations}, we show the average log likelihood on the five shapes of SYMSOL I through ablations to various aspects of the method.
The top row shows the dependence on the size of the dataset.
Performance levels off after 50,000 images per shape, but is greatly diminished for only 10,000 examples.
Note almost all of the values for 10,000 images are less than the log likelihood of a uniform distribution over SO(3), $-\log A = -2\log \pi = -2.29$, the `safest' distribution to output if training is unsuccessful.
This indicates overfitting: with only one rotation for each training example, a minimal number of examples is needed to connect all the ground truths with each view of a shape.
The network becomes confident about the rotations it has seen paired with a particular view, and assigns small probability to the unseen ground truths, resulting in large negative log likelihood values. 

The second row varies the positional encoding applied to the rotations when querying the MLP.
0 positional encoding terms corresponds to no positional encoding at all: the flattened rotation matrix is used as the query rotation.
The positional encoding benefits the three shapes with discrete symmetries and is neutral or even slightly negative for the cone and cylinder.
Intended to facilitate the representation of high frequency features~\cite{mildenhall2020nerf}, positional encoding helps capture the twelve modes of tetrahedral symmetry with two terms, whereas four are necessary for peak performance on the cube and icosahedron. 
For all shapes, including more positional encoding terms eventually degrades the performance.

In the third row, we compare different formats for the query rotation, pre-positional encoding.
For all shapes, representing rotations as matrices is optimal, with axis-angle and quaternion formats comparable to each other and a fair amount worse.
Representing rotations via Euler angles averages out near the log likelihood of a uniform distribution ($-2.29$), though with a large spread which indicates most but not all runs fail to train.

Finally, the fourth row examines the effect of normalization in the likelihood loss during training.
Randomly sampling queries from \SO(3) offers simplicity and freedom over the exact number of queries, but results in inexact normalization of the probability distribution.
During training, this leads to roughly equivalent performance as when an equivolumetric grid of queries is used, which can be exactly normalized.

\begin{figure}[htbp]
    \centering
    \includegraphics[width=0.75\linewidth]{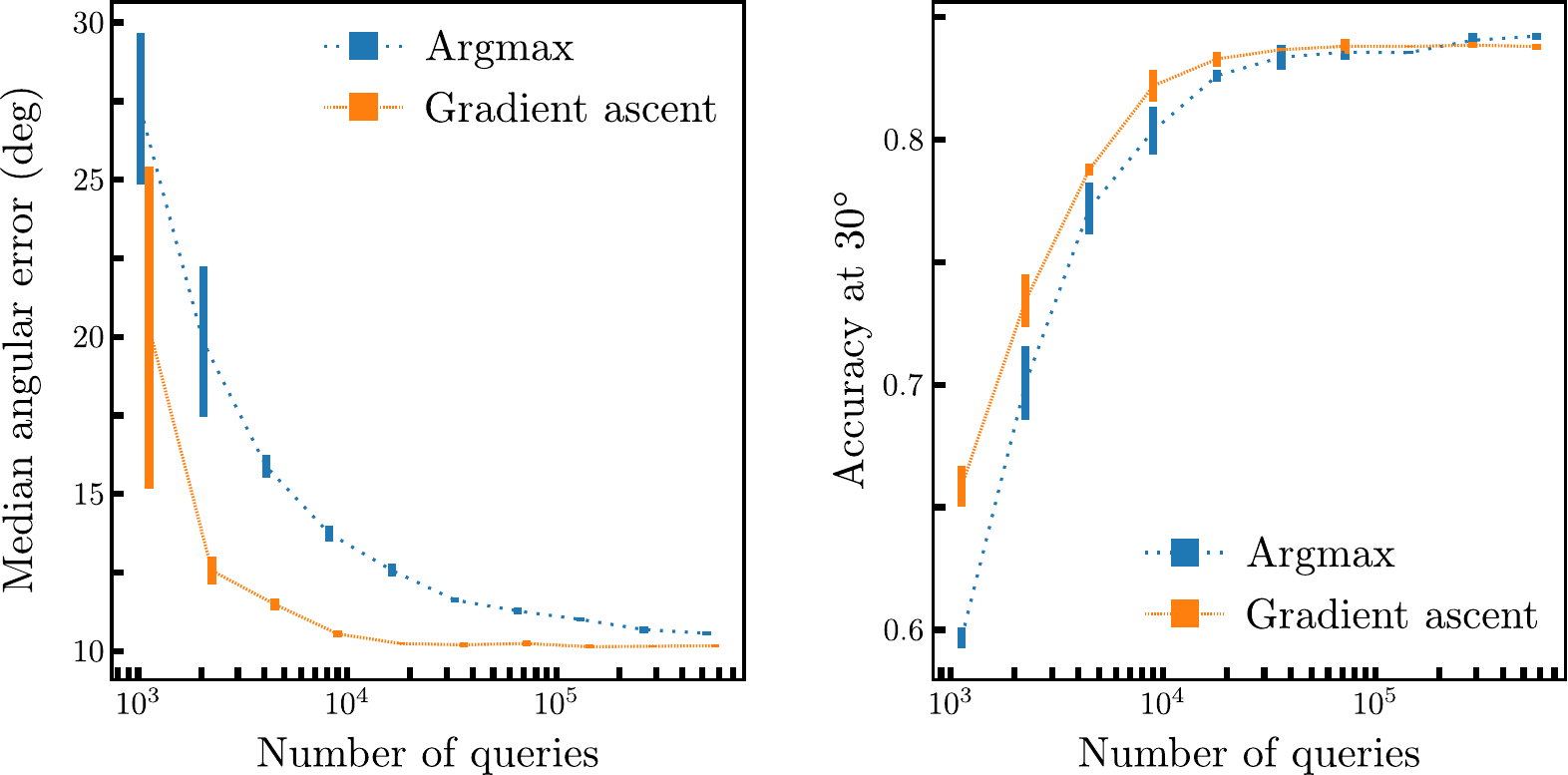}
    \caption{\textbf{The efficacy of gradient ascent on Pascal3D+.} 
    We report the average performance across classes on Pascal3D+, for the same IPDF model, using different means to extract a single-valued pose estimate.
    The error bars are the standard deviation among random sampling attempts, and the curves are slightly offset horizontally for clarity.
    }
\label{fig:supp_grad_ascent}
\end{figure}

In Figure~\ref{fig:supp_grad_ascent} we show the efficacy of performing gradient ascent to extract the most likely pose from IPDF, given an image. 
The first way to find the rotation with maximal probability is by sampling from \SO(3) and taking the \texttt{argmax} over the unnormalized outputs of IPDF.
Predictably, finer resolution of the samples yields more accurate predictions, indicated by shrinking median angular error (left) and growing accuracy at 30$^\circ$ (right) averaged over the categories of Pascal3D+.
The second way to produce an estimate leverages the fact that IPDF is fully differentiable.
We use the best guess from a sampling of queries as a starting value for gradient ascent on the output of IPDF.
The space of valid rotations is embedded in a much larger query space, so we project the updated query back to \SO(3) after every step of gradient ascent, and run it for 100 steps.
The estimates returned by gradient ascent yield optimal performance for anything more than 10,000 queries, whereas \texttt{argmax} requires more than 500,000 queries for similar results.
The difference between the \texttt{argmax} and gradient ascent is primarily in the median angular error (left): improvements of an estimate on the order of a degree would benefit this statistic more than the accuracy at 30$^\circ$.

\section{Metrics for evaluation: extended discussion}
\subsection{Prediction as a distribution: spread and average log likelihood}
\begin{figure}[htbp]
    \centering
    \includegraphics[width=0.9\linewidth]{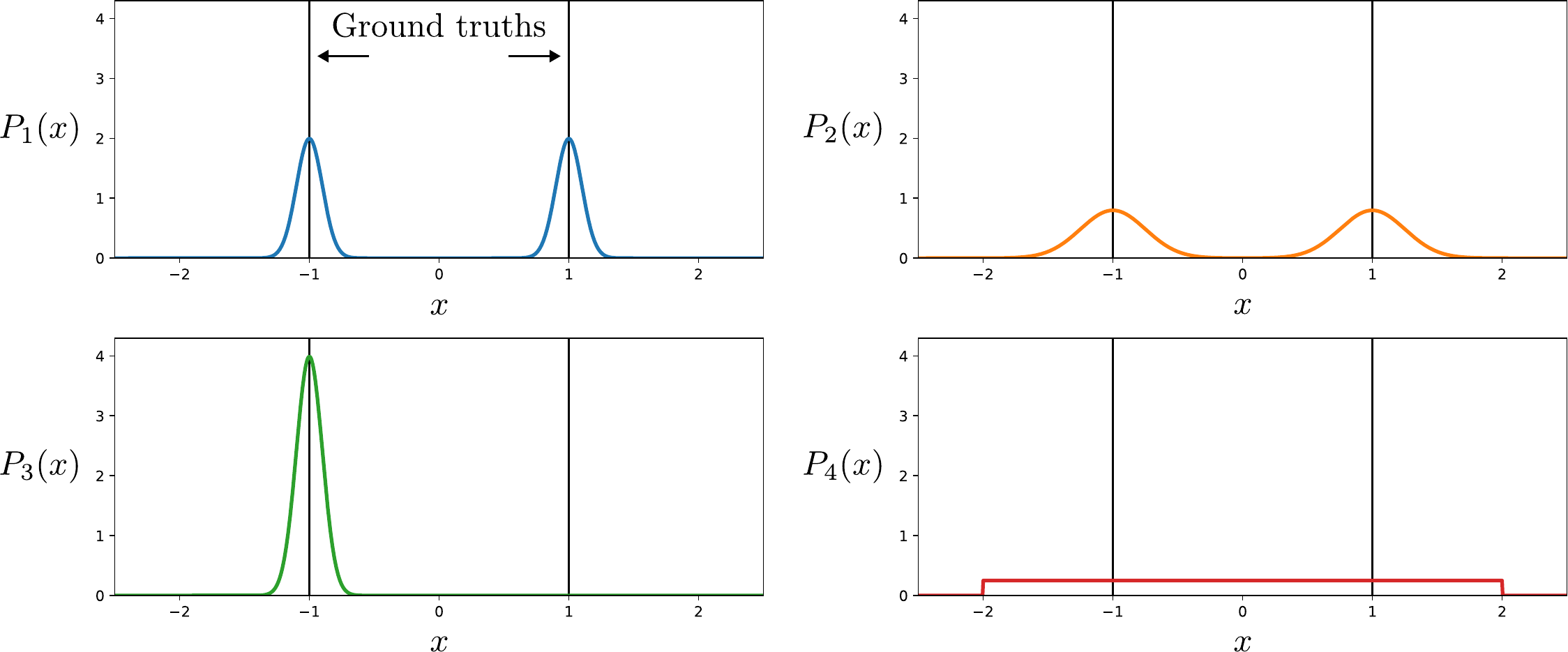}
    \caption{\textbf{Distributions modelling a scenario with multiple ground truths.} 
    $P_1$ and $P_2$ are mixtures of two normal distributions, with the components centered on the ground truths at $x=\pm 1$.
    $P_3$ is a normal distribution centered on only one of the two ground truths.
    $P_4$ is a uniform distribution over the interval $[-2, 2]$.}
\label{fig:supp_toy}
\end{figure}

Here we compare the metrics used in the main text on a simplified example in one dimension, where the ground truth consists of two values: $\{x_{GT}\}=\pm 1$.
We evaluate the four distributions ($P_1, P_2, P_3, P_4)$) shown in Figure~\ref{fig:supp_toy} which model the ground truth to varying degree. 

\begin{table}[htbp]
  \caption{Distribution-based evaluation metrics from the main text.
    }
  \centering
  {
    \begin{tabular}{@{}l
      S[table-format=1.2,table-figures-decimal=2,table-auto-round]
      S[table-format=1.2,table-figures-decimal=2,table-auto-round]
      S[table-format=1.2,table-figures-decimal=2,table-auto-round]
      S
      S[table-format=1.2,table-figures-decimal=2,table-auto-round]
      S[table-format=1.2,table-figures-decimal=2,table-auto-round]
      }
      \toprule
      & \multicolumn{2}{c}{Full GT at evaluation}  & & \multicolumn{2}{c}{Partial GT at evaluation}  \\                              
      \cmidrule{2-3} \cmidrule{5-6}
                  Distribution                     & {Spread $\downarrow$} & {Average log likelihood $\uparrow$}  &  & {Spread $\downarrow$} & {Average log likelihood $\uparrow$}    \\
      \midrule
                  $P_1=\frac{1}{2}(\mathcal{N}(-1, 0.1^2) + \mathcal{N}(1, 0.1^2))$   & \bgd 0.08 & \bgd 0.69 & & 1.04  & \bgd 0.69 & \\
                  $P_2=\frac{1}{2}(\mathcal{N}(-1, 0.25^2) + \mathcal{N}(1, 0.25^2))$  & 0.199 & \bgl -0.23 & & 1.10   & \bgl -0.23 & \\
                  $P_3=\mathcal{N}(-1, 0.1^2)$                                      & \bgd 0.08  & -98.62 & & 1.04   & -98.62 & \\
                  $P_4=\mathcal{U}(-2, 2)$                                           &  0.50  & -1.39 & & 1.25  & -1.39 & \\
      \bottomrule
    \end{tabular}
    
  }
\label{tab:supp_toy}
\end{table}

The results for the spread and average log likelihood, defined in the main text, are shown in Table~\ref{tab:supp_toy}.
There are several takeaways from this simplified example.
The spread, being the average over the ground truths of the minimum error, captures how well \textit{any} of the ground truths are represented.
By this metric, $P_1$ and $P_3$ are equivalent.
When the full set of ground truths is not known at evaluation, the spread ceases to be meaningful.

The average log likelihood measures how well \textit{all} ground truths are represented and is invariant to whether the full set of GTs is provided with each test example, or only a subset.
The latter is the predominant scenario for pose estimation datasets, where annotations are not provided for near or exact symmetries.
This means only one ground truth is provided for each test example, out of possibly several equivalent values.
In Table~\ref{tab:supp_toy}, the average log likelihood ranks the distributions in the order one would expect, with the `ignorant' uniform distribution ($P_4$) performing slightly worse than $P_1$ and $P_2$, and with $P_3$ severely penalized for failing to cover both of the ground truths.

\subsection{Prediction as a finite set and unknown symmetries: top-$k$}
\label{sec:topk}
For the case where only a single ground truth is available, despite potential symmetries,
the log-likelihood metric is the only one that is still meaningful unchanged.

Precision and spread metrics are misleading because
they penalize correct predictions that don't have a corresponding annotation.
Our solution is to drop the precision metric
and split the distribution into different modes to compute the spreads,
by finding connected components in probability distribution predicted.

The recall metrics are problematic when viewed independently of precision,
since they can be easily optimized for by returning a large number of candidate poses
covering the whole space.
Our solution here is to limit the number of output pose candidates to $k$,
yielding metrics that we denote the
top-$k$ accuracy@15\textdegree, top-$k$ accuracy@30\textdegree, and top-$k$ error.
For example, the metrics reported by \citet{s2reg,mohlin2020probabilistic} on ModelNet10-SO(3)
are equivalent to our top-1.

One issue with the top-$k$ evaluation is that we cannot disentangle if errors
are due to the dataset (lack of symmetry annotations), or due to the model.
Since there is no way around it without expensive annotation,
we find it useful to report the top-$k$ for different $k$, including $k=1$,
where no model errors are forgiven. 

Now, for each entry in the dataset,
$R_{GT}$ is the single annotated ground truth, 
the top-$k$ pose predictions are $\{\hat{R}_i\}_{1 \le i \le k}$,
and we have $k$ normalized probability distributions corresponding to each of the top-$k$ modes,
$\{\hat{p}_i\}_{1 \le i \le k}$.
The following equations describe the metrics,
\begin{align}
  \text{top-$k$ accuracy@$\alpha$} &= 
  \left[\min_{1 \le j \le k}\left\{d(R_{GT}, \hat{R_j})\right\} < \alpha \right], \\  
  \text{top-$k$ error} &=
  \min_{1 \le j \le k}d(R_{GT}, \hat{R_j}), \\
  \text{top-$k$ spread} &=
  \min_{1 \le j \le k}\left\{\int_{\SO(3)} \hat{p}_j(R) d(R, R_{GT})\,dR\right\}.
\end{align}

Typically, accuracy and spread are averaged over the whole dataset,
while the median error over all entries is reported.

\section{ModelNet10-SO(3) detailed results}
\label{sec:supp_m10}

\Cref{tab:m10so3} extends the ModelNet10-SO(3) table in the main paper and shows per-category metrics.

Since our model predicts a full distribution of rotations,
we find the modes of this distribution,
by first thresholding by density and then assigning to the same mode any two points
that are closer than a second threshold.
This method outputs a variable number of modes for each input,
as opposed to methods based on mixtures of unimodal distributions~\cite{gilitschenski2019deep,deng2020deep},
where the number of modes is a fixed hyperparameter.

We then rank the modes by their total probability mass,
assign their most likely pose as the mode center,
and return the top-$k$ centers for a given $k$.  
The evaluation takes the minimum error over the list of candidates,
as described in \cref{sec:topk}. 
This kind of top-$k$ evaluation is common practice for image classification tasks
like ImageNet~\cite{russakovsky2015imagenet}.

As expected, all metrics improve by increasing $k$,
but the symmetric categories, where the single ground-truth evaluation is inappropriate,
improve dramatically, suggesting that the lower top-1 performance
can indeed be attributed to the lack of symmetry annotations for evaluation
and is not a limitation of our model.

\begin{table*}[ht]
  \centering
  \resizebox{0.9\textwidth}{!}{
    \begin{tabular}{@{}ll
      ccccccccccc
      }
      \toprule
                                                           &                                      & {avg.}     & {bathtub}  & {bed}      & {chair}    & {desk}     & {dresser}  & {tv}       & {n. stand} & {sofa}     & {table}    & {toilet}   \\
      \midrule
      \multirow{4}{*}{Acc@15\textdegree}                   & \citet{deng2020deep}            & 0.562      & 0.140      & 0.788      & 0.800      & 0.345      & 0.563      & 0.708      & 0.279      & 0.733      & 0.440      & 0.832      \\      
                                                           & \citet{prokudin2018deep}        & 0.456      & 0.114      & 0.822      & 0.662      & 0.023      & 0.406      & 0.704      & 0.187      & 0.590      & 0.108      & \bgl 0.946 \\
                                                           & \citet{mohlin2020probabilistic} & \bgl 0.693 & \bgl 0.322 & \bgd 0.882 & \bgd 0.881 & \bgl 0.536 & \bgl 0.682 & \bgl 0.790 & \bgl 0.516 & \bgd 0.919 & \bgl 0.446 & \bgd 0.957 \\      
                                                           & IPDF (ours)                          & \bgd 0.719 & \bgd 0.392 & \bgl 0.877 & \bgl 0.874 & \bgd 0.615 & \bgd 0.687 & \bgd 0.799 & \bgd 0.567 & \bgl 0.914 & \bgd 0.523 & 0.945      \\
      \cmidrule{3-13}
                                                           & IPDF (ours), top-2                          & 0.868      & 0.735      & 0.946      & 0.900      & 0.803      & 0.810      & 0.883      & 0.756      & 0.959      & 0.932      & 0.960      \\
                                                           & IPDF (ours), top-4                          & 0.904      & 0.806      & 0.966      & 0.905      & 0.862      & 0.870      & 0.899      & 0.842      & 0.966      & 0.956      & 0.963      \\
      \midrule
      \multirow{4}{*}{Acc@30\textdegree}                   & \citet{deng2020deep}            & 0.694      & 0.325      & 0.880      & 0.908      & 0.556      & 0.649      & 0.807      & 0.466      & 0.902      & 0.485      & 0.958      \\      
                                                           & \citet{prokudin2018deep}        & 0.528      & 0.175      & 0.847      & 0.777      & 0.061      & 0.500      & 0.788      & 0.306      & 0.673      & 0.183      & \bgl 0.972 \\
                                                           & \citet{mohlin2020probabilistic} & \bgd 0.757 & \bgl 0.403 & \bgd 0.908 & \bgd 0.935 & \bgd 0.674 & \bgd 0.739 & \bgd 0.863 & \bgd 0.614 & \bgd 0.944 & \bgl 0.511 & \bgd 0.981 \\      
                                                           & IPDF (ours)                          & \bgl 0.735 & \bgd 0.410 & \bgl 0.883 & \bgl 0.917 & \bgl 0.629 & \bgl 0.688 & \bgl 0.832 & \bgl 0.570 & \bgl 0.921 & \bgd 0.531 & 0.967      \\
      \cmidrule{3-13}
                                                           & IPDF (ours), top-2                          & 0.888      & 0.770      & 0.953      & 0.946      & 0.825      & 0.812      & 0.918      & 0.762      & 0.968      & 0.945      & 0.982      \\
                                                           & IPDF (ours), top-4                          & 0.926      & 0.846      & 0.973      & 0.953      & 0.889      & 0.874      & 0.939      & 0.851      & 0.975      & 0.972      & 0.988      \\
      \midrule
      \multirow{4}{*}{\shortstack{Median \\ Error ($^\circ$)}} & \citet{deng2020deep}            & 32.6       & 147.8      & 9.2        & 8.3        & 25.0       & 11.9       & 9.8        & 36.9       & 10.0       & 58.6       & 8.5        \\
                                                           & \citet{prokudin2018deep}        & 49.3       & \bgl 122.8 & \bgd 3.6   & 9.6        & 117.2      & 29.9       & 6.7        & 73.0       & 10.4       & 115.5      & \bgl 4.1   \\
                                                           & \citet{mohlin2020probabilistic} & \bgd 17.1  & \bgd 89.1  & \bgl 4.4   & \bgd 5.2   & \bgl 13.0  & \bgl 6.3   & \bgl 5.8   & \bgl 13.5  & \bgd 4.0   & \bgl 25.8  & \bgd 4.0   \\      
                                                           & IPDF (ours)                          & \bgl 21.5  & 161.0      & \bgl 4.4   & \bgl 5.5   & \bgd 7.1   & \bgd 5.5   & \bgd 5.7   & \bgd 7.5   & \bgl 4.1   & \bgd 9.0   & 4.8        \\
      \cmidrule{3-13}
                                                           & IPDF (ours), top-2                          & 4.9     & 6.8     & 4.1     & 5.5     & 5.3     & 4.9     & 5.3     & 5.1     & 3.9     & 3.7     & 4.8     \\
                                                           & IPDF (ours), top-4                          & 4.8     & 6.0     & 4.1     & 5.4     & 5.1     & 4.7     & 5.2     & 4.8     & 3.9     & 3.7     & 4.8 \\
      \bottomrule
    \end{tabular}
  }
  \caption{ModelNet10-SO(3) per-category results.
  }
  \label{tab:m10so3}
\end{table*}

\section{Implementation specifics}
We train with the Adam optimizer ($\beta_1 = 0.9$, $\beta_2 = 0.999$) with a linear warm up to the base learning rate of $10^{-4}$ over 1000 steps, and then cosine decay to zero over the remainder of training.  
\begin{table*}[h]
{
\begin{center}
\small
\begin{tabular}{cccc}
Layer &  Activation & Output \\
\hline \\
Vision Description Input & - & 2048 \\
Rotation Input & - & [3, 3] \\
Flatten & - & 9 \\
Positional Encoding & - & [$2m \times $9] \\
Concatenate & - & [2048 + $2m \times $9] \\
Dense & ReLU & 256 & \\
& & \dots & $\times n$\\
Dense & None & 1 \\
\end{tabular}
\end{center}
}
\caption{\textbf{IPDF architecture.} $m$ is the number of positional encoding frequencies and $n$ is the number of fully connected layers in the MLP.  The factor of 2 comes from using both sines and cosines in the positional encoding. 
The vision description is the result of applying global average pooling to the output of an ImageNet pre-trained ResNet to obtain a 2048-dimensional vector.
We use an ImageNet pre-trained Resnet50 for SYMSOL, T-LESS, and ModelNet10-SO(3), and Resnet101 for Pascal3D+. }
\label{tab:arch}
\end{table*}

\paragraph{Efficient implementation}
The input to the MLP is a concatenation of the image descriptor produced by a CNN and a query pose. During both training and inference, we evaluate densities for a large number of poses per image.
A naive implementation would replicate and tile image descriptors $\{d_i\}_{0 \leq i < N_B}$ 
and pose queries $\{q_j\}_{0 \leq j < N_Q}$, where $N_B$ is the mini-batch size and $N_Q$ is the
number of pose queries, and evaluate the first fully connected operation with weights $W$
(before applying bias and nonlinearity)
in a batched fashion, as follows,
\begin{align}
  \label{eq:slow}
  W
  \begin{bmatrix}
    d_1 & d_1 & d_1 & \cdots & d_2 & d_2 & d_2 & \cdots \\
    q_1 & q_2 & q_3 & \cdots & q_1 & q_2 & q_3 & \cdots
  \end{bmatrix}.
\end{align}
When computed this way, this single step is the computational bottleneck. 
An alternative, much more efficient method is to observe that
\begin{align}
  W \twobyone{d_i}{q_j} = W\twobyone{d_i}{0} + W\twobyone{0}{q_j} = W_d d_i + W_qq_j,
\end{align}
where $W = [W_d \quad W_q]$.
In this manner, $W_d$ can be applied batchwise to image descriptors,
yielding a $N_O \times N_B$ output,
and $W_q$ can be applied to all query poses independently,
yielding a $N_O \times N_Q$ output,
where $N_O$ is the number of output channels (number of rows in $W$). 
An $N_O \times N_Q \times N_B$ tensor equivalent to \cref{eq:slow} is then obtained
via a broadcasting sum, drastically reducing the number of operations.

\paragraph{SYMSOL}
For the SYMSOL experiments, three positional encoding terms were used for the query, and four fully connected layers of 256 units with ReLU activation for the MLP.
One network was trained for all five shapes of SYMSOL I with a batch size of 128 images for 100,000 steps (28 epochs).  
A different network was trained for each of the three textured shapes of SYMSOL II; these trained with a batch size of 64 images for 50,000 steps (36 epochs).
The loss calculation requires evaluating a coverage of points on SO(3) along with the ground truth in order to find the approximate normalization rescaling of the likelihoods.
We found that this coverage did not need to be particularly dense, and used 4096 points for training.

\paragraph{T-LESS}
For T-LESS, only one positional encoding term was used, and the MLP consisted of a single layer of 256 units with ReLU activation.  
The images were color-normalized and tight-cropped as in \citet{gilitschenski2019deep}.
Training was with a batch size of 64 images for 50,000 steps (119 epochs).

\paragraph{ModelNet10-SO(3)}
For ModelNet10-SO(3)~\cite{s2reg},
we use four fully connected layers of 256 units with ReLU activation as in SYMSOL.
We train a single model for the whole dataset, for 100,000 steps with batch size of 64.
Following \citet{s2reg} and \citet{mohlin2020probabilistic}, we concatenate a one-hot encoding of the
class label to the image descriptor before feeding it to the MLP.

\paragraph{Pascal3D+}
We used a learning rate of $10^{-5}$ for 150,000 steps, with the same schedule as in the other experiments (linear ramp for the first 1000 steps, then cosine decay).  The vision model was an ImageNet pre-trained ResNet101, and the MLP consisted of two fully connected layers of 256 units with ReLU activation (trained on all classes at once, without class label information).
We supplemented the Pascal3D+ training images with synthetic images from Render for CNN~\cite{su2015render}, such that every mini-batch of 64 images consisted of 25\% real images and 75\% synthetic.

\subsection{Baseline methods}
\paragraph{[\citet{deng2020deep}]}
We trained the multi-modal Bingham distribution model from \citet{deng2020deep} using their PyTorch code.\footnote{\url{https://github.com/Multimodal3DVision/torch\_bingham}.} Note, this is a follow-up work of an earlier paper which references the same implementation~\citep{deng2020deep}. Our only modification was a minor one to remove the translation component from the model as only the rotation representation needs to be learned. We found the model performed best with the same general settings as used in the reference paper (rWTA loss with two stage training -- first stage trains rotations only, the second stage trains both rotations and mixture coefficients). 

For the ModelNet10-SO(3) and SYMSOL datasets we trained a single model per shape category, 
and we found no benefit with increasing the number of components (we used 10 for ModelNet10 and 16 for SYMSOL).

\paragraph{[\citet{gilitschenski2019deep}]}
We trained the multi-modal Bingham distribution model from \citet{gilitschenski2019deep} using their PyTorch code.\footnote{\url{https://github.com/igilitschenski/deep\_bingham}.} For this baseline we again trained a single model per shape for ModelNet10-SO(3) and SYMSOL. We followed the published approach and trained the model in two stages -- first stage with fixed dispersion and second stage updates all distribution parameters. For a batch size of 32, a single training step for a 4-component distribution takes almost 2 seconds on a NVIDIA TESLA P100 GPU. The time is dominated by the lookup table interpolation to calculate the distribution's normalizing term (and gradient), and is linear in the number of mixture components (training with 12 mixture components took over 7 seconds per step). This limited our ability to tune hyperparameters effectively or train with a large number of mixture components.

\paragraph{[\citet{prokudin2018deep}]}
We trained the infinite mixture model from \citet{prokudin2018deep} using their Tensorflow code.\footnote{\url{https://github.com/sergeyprokudin/deep_direct_stat}.}
The only modification was during evaluation: the log likelihood required our method of normalization via equivolumetric grid because representing a distribution over ${\SO(3)}$ as the product of three individually normalized von Mises distributions lacks the necessary Jacobian.
We left the improperly normalized log likelihood in their loss, as it was originally formulated.
A different model was trained per shape category of SYMSOL and ModelNet10-SO(3). 

Note that our  implicit pose distribution is trained as a single model
for the whole of SYMSOL I and ModelNet10-SO(3) datasets,
so the comparisons against \citet{deng2020deep}, \citet{gilitschenski2019deep}, and \citet{prokudin2018deep}
favor the baselines.
Our method outperforms them nevertheless.

\subsection{A note on Pascal3D+ evaluations with respect to \citeauthor{s2reg} and \citeauthor{mohlin2020probabilistic}}
In the Pascal3D+ table in the main paper, and mentioned in that caption, we report numbers for \citet{s2reg} and \citet{mohlin2020probabilistic} which differ from the numbers reported in their papers (these are the rows marked with \ddag).
\paragraph{\citet{s2reg}}
An error in the evaluation code, reported on github\footnote{\url{https://github.com/leoshine/Spherical\_Regression/issues/8}}, incorrectly measured the angular error -- reported numbers were incorrectly lower by a factor of $\sqrt 2$. The authors corrected the evaluation code for ModelNet10-SO(3) and posted updated numbers, which we show in our paper. However, their evaluation code used for Pascal3D+ still contains the incorrect $\sqrt 2$ factor: comparing the corrected ModelNet10-SO(3) geodesic distance function\footnote{\url{https://github.com/leoshine/Spherical\_Regression/blob/a941c732927237a2c7065695335ed949e0163922/S3.3D\_Rotation/lib/eval/GTbox/eval\_quat\_multilevel.py\#L45}} and the Pascal3D+ geodesic distance function\footnote{\url{https://github.com/leoshine/Spherical\_Regression/blob/a941c732927237a2c7065695335ed949e0163922/S1.Viewpoint/lib/eval/eval\_aet\_multilevel.py\#L135}} the $\sqrt 2$ difference is clear. We sanity checked this by running their Pascal3D+ code with the incorrect metric and were able to closely match the numbers in the paper. In the main paper, we report performance obtained using the corrected evaluation code.

\paragraph{\citet{mohlin2020probabilistic}}
We found that the code released by \cite{mohlin2020probabilistic} uses different dataset splits for training and testing on Pascal3D+ than many of the other baselines we compared against. Annotated images in the Pascal3D+ dataset are selected from one of four source image sets: ImageNet\_train, ImageNet\_val, PASCALVOC\_train, and PASCALVOC\_val. Methods like \citeauthor{mahendranmixed} and \citeauthor{s2reg} place all the ImageNet images (ImageNet\_train, ImageNet\_val) in the training partition (i.e.\ used for training and/or validation): \textit{``We use the ImageNet-trainval and Pascal-train images as our training data and the Pascal-val images as our testing data.''} \citet{mahendranmixed}, Sec 4. However, in the code released by~\citet{mohlin2020probabilistic}, we observe the test set is sourced from the ImageNet data\footnote{\url{https://github.com/Davmo049/Public\_prob\_orientation\_estimation\_with\_matrix\_fisher\_distributions/blob/4baba6d06ca36db4d4cf8c905c5c3b70ab5fb54a/Pascal3D/Pascal3D.py\#L558-L583}}.  We reran the \citeauthor{mohlin2020probabilistic} code as-is and were able to match their published numbers. After logging both evaluation loops, we confirmed the test data differs between \citeauthor{mohlin2020probabilistic} and \citeauthor{s2reg}. The numbers we report in the main paper for \citeauthor{mohlin2020probabilistic} are after modifying the data pipeline to match \citeauthor{s2reg}, which is also what we follow for our IPDF experiments. We ran \citeauthor{mohlin2020probabilistic} with and without augmentation and warping in the data pipeline and chose the best results (which was with warping and augmentation).

\end{document}